\documentclass[10pt,twocolumn,letterpaper]{article}

\usepackage[pagenumbers]{cvpr}              % To produce the CAMERA-READY version
\usepackage{dsfont}
\usepackage{graphicx}
\usepackage{amsmath}
\usepackage{amssymb}
\usepackage{booktabs}

\usepackage{listings}
\usepackage{fancyvrb}

\usepackage{colortbl}
\usepackage{pgf}
\usepackage{pgfplotstable}
\usepackage{array}
\usepackage{csquotes}

\definecolor{customblack}{RGB}{105, 105, 105} % Soft black (Dim Gray)
\newcommand{\applycolorblack}[1]{
    \pgfmathparse{int(#1*100)} % Multiply #1 by 100 and cast to an integer
    \ifnum\pgfmathresult>70 \cellcolor{customblack!80}\else
    \ifnum\pgfmathresult>60 \cellcolor{customblack!70}\else
    \ifnum\pgfmathresult>50 \cellcolor{customblack!60}\else
    \ifnum\pgfmathresult>40 \cellcolor{customblack!50}\else
    \ifnum\pgfmathresult>30 \cellcolor{customblack!40}\else
    \ifnum\pgfmathresult>20 \cellcolor{customblack!30}\else
    \ifnum\pgfmathresult>10 \cellcolor{customblack!20}\else
    \ifnum\pgfmathresult>0 \cellcolor{customblack!10}\fi\fi\fi\fi\fi\fi\fi\fi
    #1
}

\definecolor{customred}{RGB}{255, 182, 193} % Light Pink
\newcommand{\applycolorred}[1]{
    \pgfmathparse{int(#1*-100)} % Multiply #1 by 100 and cast to an integer
    \ifnum\pgfmathresult>70 \cellcolor{customred!80}\else
    \ifnum\pgfmathresult>60 \cellcolor{customred!70}\else
    \ifnum\pgfmathresult>50 \cellcolor{customred!60}\else
    \ifnum\pgfmathresult>40 \cellcolor{customred!50}\else
    \ifnum\pgfmathresult>30 \cellcolor{customred!40}\else
    \ifnum\pgfmathresult>20 \cellcolor{customred!30}\else
    \ifnum\pgfmathresult>10 \cellcolor{customred!20}\else
    \ifnum\pgfmathresult>0 \cellcolor{customred!10}\fi\fi\fi\fi\fi\fi\fi\fi
    #1
}

\definecolor{customblue}{RGB}{173, 216, 230}
\newcommand{\applycolorblue}[1]{
    \pgfmathparse{int(#1*100)} % Multiply #1 by 10 and cast to an integer
    \ifnum\pgfmathresult>70 \cellcolor{customblue!80}\else
    \ifnum\pgfmathresult>60 \cellcolor{customblue!70}\else
    \ifnum\pgfmathresult>50 \cellcolor{customblue!60}\else
    \ifnum\pgfmathresult>40 \cellcolor{customblue!50}\else
    \ifnum\pgfmathresult>30 \cellcolor{customblue!40}\else
    \ifnum\pgfmathresult>20 \cellcolor{customblue!30}\else
    \ifnum\pgfmathresult>10 \cellcolor{customblue!20}\else
    \ifnum\pgfmathresult>0 \cellcolor{customblue!10}\fi\fi\fi\fi\fi\fi\fi\fi
    #1
}

\usepackage{multirow}
\usepackage{graphicx}
\usepackage{rotating}

\usepackage{pifont}
\usepackage{xcolor}

\newcommand{\bcmark}{\ding{51}} % black checkmark
\newcommand{\bxmark}{\ding{55}}   % black crossmark

\usepackage{enumitem}

\newcommand{\para}[1]{\medskip\noindent\textbf{#1.}}

\newcommand{\framework}{PARC}

\definecolor{promptcol}{rgb}{0.66,0.38,0.14}
\definecolor{reliabcol}{rgb}{0.14,0.32,0.56}
\definecolor{calibcol}{rgb}{0.41,0.13,0.27}
\definecolor{questcol}{rgb}{0.27,0.36,0.15}
\definecolor{plotblue}{rgb}{0.26,0.51,0.85}
\definecolor{plotorange}{rgb}{0.94,0.6,0.53}

\definecolor{reliabred}{rgb}{0.59,0.16,0.09}
\definecolor{reliabblue}{rgb}{0.00,0.22,0.50}

\definecolor{cvprblue}{rgb}{0.21,0.49,0.74}
\usepackage[pagebackref,breaklinks,colorlinks,allcolors=cvprblue]{hyperref}

\usepackage[capitalize]{cleveref}
\crefname{section}{Sec.}{Secs.}
\Crefname{section}{Section}{Sections}
\Crefname{table}{Table}{Tables}
\crefname{table}{Tab.}{Tabs.}

\def\paperID{3317} % *** Enter the Paper ID here
\def\confName{CVPR}
\def\confYear{2025}

\title{\framework{}: A Quantitative Framework\\Uncovering the Symmetries within Vision Language Models}

\author{
Jenny Schmalfuss\renewcommand{\thefootnote}{\arabic{footnote}}\footnotemark[1]
\hspace*{.025cm}\renewcommand{\thefootnote}{\fnsymbol{footnote}}\footnotemark[2]
\hspace*{.1cm}
Nadine Chang\renewcommand{\thefootnote}{\arabic{footnote}}\footnotemark[2]
\hspace*{.025cm}\renewcommand{\thefootnote}{\fnsymbol{footnote}}\footnotemark[1]
\hspace*{.1cm}
Vibashan VS\renewcommand{\thefootnote}{\arabic{footnote}}\footnotemark[3]
\hspace*{.05cm}\renewcommand{\thefootnote}{\fnsymbol{footnote}}\footnotemark[2]
\hspace*{.1cm}
Maying Shen\renewcommand{\thefootnote}{\arabic{footnote}}\footnotemark[2]
\hspace*{.1cm}
Andrés Bruhn\renewcommand{\thefootnote}{\arabic{footnote}}\footnotemark[1]
\hspace*{.1cm}
Jose M. Alvarez\renewcommand{\thefootnote}{\arabic{footnote}}\footnotemark[2]
\\
\renewcommand{\thefootnote}{\arabic{footnote}}\footnotemark[1]
\hspace*{.05cm}University of Stuttgart
\hspace*{.3cm}
\renewcommand{\thefootnote}{\arabic{footnote}}\footnotemark[2]
\hspace*{.05cm}NVIDIA
\hspace*{.3cm}
\renewcommand{\thefootnote}{\arabic{footnote}}\footnotemark[3]
\hspace*{.05cm}Johns Hopkins University
}

\begin{document}

\twocolumn[{%
\renewcommand\twocolumn[1][]{#1}%
\maketitle
\vspace{-1.cm}
\begin{center}
    \includegraphics[width=\linewidth]{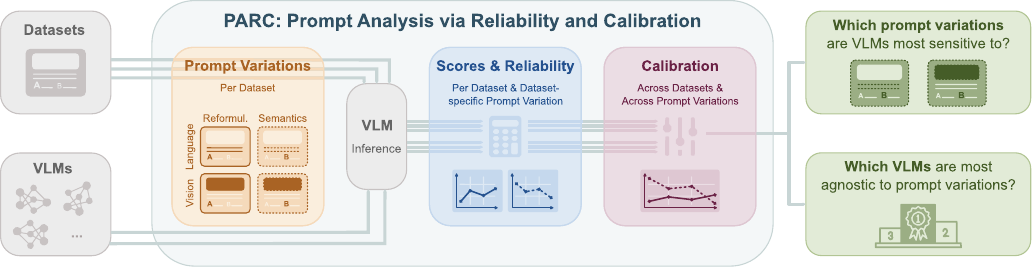}
    \vspace{-0.6cm}
    \captionof{figure}{\framework{} prompt sensitivity analysis framework overview. Given a collection of VLMs and datasets, \framework{} identifies \textcolor{questcol}{which \textbf{prompt variations} these VLMs are most sensitive to}, and \textcolor{questcol}{which \textbf{VLMs} are most agnostic to prompt variations \emph{[green]}}. To achieve this, \framework{} first applies systematic \textcolor{promptcol}{\textbf{prompt variations}} \textcolor{promptcol}{\emph{[orange]}} to the language and vision components of the datasets, then evaluates the VLM performance on these varied datasets with multiple established scores and a novel \textcolor{reliabcol}{\textbf{reliability score}} \textcolor{reliabcol}{\emph{[blue]}}, and finally \textcolor{calibcol}{\textbf{calibrates}} \textcolor{calibcol}{\emph{[red]}} those scores to make them directly comparable across the diverse input datasets as well as \framework{}'s prompt variations. \label{fig:teaser}}
\end{center}
}]

\renewcommand{\thefootnote}{\fnsymbol{footnote}}
\footnotetext[2]{Work done during an internship at NVIDIA.}
% \begingroup
% \renewcommand\thefootnote{}\footnote{JS acknowledges IMPRS-IS.}
% \endgroup
\footnotetext[1]{Corresponding author. nadinec@nvidia.com}
\renewcommand{\thefootnote}{\arabic{footnote}}

\begin{abstract}
Vision language models (VLMs) respond to user-crafted text prompts and visual inputs, and are applied to numerous real-world problems.
VLMs integrate visual modalities with large language models (LLMs), which are well known to be prompt-sensitive.
Hence, it is crucial to determine whether VLMs inherit this instability to varying prompts.
We therefore investigate \emph{which prompt variations VLMs are most sensitive to} and \emph{which VLMs are most agnostic to prompt variations}.
To this end, we introduce \framework{} (\underline{P}rompt \underline{A}nalysis via \underline{R}eliability and \underline{C}alibration), a VLM prompt sensitivity analysis framework built on three pillars: (1) plausible prompt variations in both the language and vision domain, (2) a novel model reliability score with built-in guarantees, and (3) a calibration step that enables dataset- and prompt-spanning prompt variation analysis.
%\todo{Stronger take home, answer questions}
% Regarding prompt variations, experimental results from \framework{} show that VLMs mirror LLM language prompt sensitivity in the vision domain, and most destructive variations are those that change the expected answer. 
Regarding prompt variations, \framework{}'s evaluation shows that VLMs mirror LLM language prompt sensitivity in the vision domain, and most destructive variations change the expected answer. 
% Regarding models, InternVL2 models are outstandingly robust among 22 evaluated VLMs, which links prompt sensitivity to training data.
Regarding models, outstandingly robust VLMs among 22 evaluated models come from the InternVL2 family.
We further find indications that prompt sensitivity is linked to training data.
% Intruigingly, prompt sensitivity is likely linked to training data.
\hyperlink{https://github.com/NVlabs/PARC}{https://github.com/NVlabs/PARC}
% Code and datasets will be released.
%\todo{publicly release the \framework{} framework + mention in Intro, conclusion}
%We further find significant reliability differences among VLM families, with InternVL2 as most robust, and find indications that prompt sensitivity is more closely linked to training data than model size.
%
%across datasets that prompt robustness is similar in VLM families, meaning models for safety-critical tasks should be selected by family and not model size alone.
\end{abstract}
\vspace{-1.5\baselineskip}
\section{Introduction}
\vspace{-.25\baselineskip}

Vision language models (VLMs) are being adopted in safety-critical applications such as autonomous navigation~\cite{shah2023robotnav,li2024hydra} or disease screening~\cite{qinmedical,royer2024multimedeval}.
They can respond to user inputs, called \emph{prompts}, that include both text and images.
But to leverage their full practical potential, VLMs need to be reliable:
When presented with two roadside images and asked \enquote{Which lane has people?}, VLMs should respond with the correct image regardless of image arrangement. 
However, we find that current VLMs deliver inconsistent results across such input changes.
While this phenomenon, called \textit{prompt sensitivity}, receives increasing attention for Large Language Models (LLMs)~\cite{elazar2021measuring,jang-etal-2022-becel,gan-mori-2023-sensitivity,mizrahi2023state,sclar2024quantifying,voronov2024mind,anagnostidis2024how}, it is hardly studied in VLMs~\cite{qi2023limitation,khan2024consistency}.

For practical applications we want to select VLMs that are agnostic to prompt variations, \ie that perform well independent of the specific way they are prompted.
Hence in this work, we critically examine the VLM sensitivity to varying user prompts.
Our investigation is driven by two guiding questions:
\emph{Which prompt variations are VLMs most sensitive to?}\ and \emph{Which VLM class is most agnostic to prompt variations?}
To answer these questions, we argue that three components are currently missing:
\begin{enumerate}[label=(\roman*),topsep=0\baselineskip]
\setlength{\itemsep}{.25\baselineskip}%
\setlength{\parskip}{0pt}%
\setlength{\topsep}{0pt}
\setlength{\labelindent}{0pt}%
    \item Realistic \textit{prompt variations} for VLMS,
    \item A notion of \textit{model reliability}, and
    \item \textit{Comparable metrics} across prompts and datasets.
\end{enumerate}
Firstly, prior work on VLM prompt sensitivity is exclusively focused on noisy, non-comprehensible prompts~\cite{khan2024consistency,qi2023limitation}, which do not reflect prompts that would be created by humans in practical applications.
Secondly, VLM and LLM reliability is currently expressed via multiple scores~\cite{kil2024compbench,liu2023mmbench,kostumov2024uncertainty,ye2024benchmarking,hudson2019gqa,jang-etal-2022-becel,khan2024consistency}, which complicates model selection because no single score describes model reliability in an easily interpretable manner.
Lastly, all current scores lack direct comparability across prompt variations and datasets due to changes in the expected random performance~\cite{anagnostidis2024how,lu2024mathvista,wang2024blink}, which prevents direct model comparisons across prompt variations and across datasets.

%\para{\framework{} prompt analysis framework}
We propose a VLM prompt sensitivity analysis framework called \textbf{\framework{}} (\underline{P}rompt \underline{A}nalysis via \underline{R}eliability and \underline{C}alibration) that addresses all three described limitations and allows identifying challenging prompt variations and prompt-agnostic VLMs.
\framework{} introduces realistic \textcolor{promptcol}{\textbf{prompt variations}}, a \textcolor{reliabcol}{\textbf{reliability score}} summarizing VLM performance as single number, and a \textcolor{calibcol}{\textbf{score calibration}} step for cross-dataset and cross-prompt comparisons.
\cref{fig:teaser} visualizes \framework{}, and we describe its three components below.

\para{Prompt variations}
We define prompt variations as changes to a prompt's language or vision component, and we identify two main groups of realistic variations:
\emph{Prompt reformulations} that rephrase the original prompt to preserve its meaning and expected answer~\cite{jang-etal-2022-becel,elazar2021measuring,mizrahi2023state,sun2024evaluating} and \emph{semantic prompt variations} that alter the original prompt meaning along with its expected answer, \ie through negations like \enquote{What is red?} vs. \enquote{What is not red?}~\cite{anschutz-etal-2023-correct,hosseini-etal-2021-understanding,zhang2023beyond}.
As prior work focuses on language variations for LLMs, we propose to mirror those language variation classes in the vision domain and explore parallels between language and vision prompts in VLMs for the first time.
To this end, we carefully craft a comprehensive set of 11 prompt variations for \framework{} that includes vision variations to match reformulations and semantic variations in the language domain.

\para{Reliability score}
Prompt sensitivity is commonly evaluated with the scores accuracy, certainty~\cite{kostumov2024uncertainty,ye2024benchmarking} and consistency~\cite{hudson2019gqa,jang-etal-2022-becel,khan2024consistency}, where consistency measures the agreement between two different prompt variations.
With \framework{}, we introduce a novel reliability score to act as summarizing layer on top of an accuracy and certainty measure.
It is crafted to highlight accurate, confident models and to flag models that confidently make wrong predictions.
In contrast to prior accuracy-certainty combinations~\cite{kostumov2024uncertainty}, our reliability score offers two explicit accuracy and certainty guarantees that can be read from a given score at a glance.

\para{Score calibration}
None of the previously mentioned scores are directly comparable across datasets or across prompt variations, because these may come with distinct task difficulties in the form of expected random performance: 
For an image of a red pen, the multiple choice question \enquote{Which object is red? (A) pen (B) notebook (C) cup} has one correct answer and is therefore more difficult to answer than its negation \enquote{Which object is not red?} with two correct answers.
Yet, standard evaluations often ignore this and only occasionally report random performance~\cite{anagnostidis2024how,lu2024mathvista,wang2024blink}.
To balance dataset and prompt difficulty, we introduce a principled calibration step that rescales scores relative to their expected random performance.
This calibration uniquely positions \framework{} to analyze VLM or LLM prompt sensitivity, because model statistics can be fairly computed across datasets and prompts alike.

\para{Contributions}
Within our prompt sensitivity analysis framework \framework{} we make three technical contributions: 
\begin{enumerate}[label=(\arabic*),topsep=0.25\baselineskip]
\setlength{\itemsep}{.25\baselineskip}%
\setlength{\parskip}{0pt}%
\setlength{\topsep}{0pt}
\setlength{\labelindent}{0pt}%
\item We propose a comprehensive set of 11 plausible \emph{prompt variations} for both language and vision prompts in VLMs, designed to capture realistic user prompts.
\item We design an interpretable \emph{reliability score} that intuitively combines a model's accuracy and confidence, and provides two explicit guarantees at a glance. 
\item We introduce a principled \emph{score calibration step} based on the expected random performance for accuracy, certainty, consistency and reliability, uniquely enabling comparisons across datasets and across prompts. 
\end{enumerate}
With \framework{}, we can experimentally analyze VLM prompt sensitivity for the first time and derive key insights that answer our guiding questions:
\begin{enumerate}[label=(\arabic*),topsep=0.25\baselineskip]
\setcounter{enumi}{3}
\setlength{\itemsep}{.25\baselineskip}%
\setlength{\parskip}{0pt}%
\setlength{\topsep}{0pt}
\setlength{\labelindent}{0pt}%
\item \textit{Which prompt variations are VLMs most sensitive to?}
VLMs are sensitive to language and vision variations, thus mirroring LLM prompt sensitivity.
They struggle most with semantic changes like negations, which alter the expected answer -- in both language and vision.
\item \textit{Which VLMs are most agnostic to prompt variations?}
Experimentally, we find that that models in the same family have similar prompt sensitivities. Currently, InternVL2 models are most reliable, hinting at training data protocols as path to alleviate excessive sensitivity.
\end{enumerate}

%%%%%%%%% METHOD
\section{PARC Framework}

\begin{table*}
    \centering
    \scalebox{0.79}{
    \begin{tabular}{@{}l@{\quad} c@{\quad}p{4.5cm}@{\ \ }l@{\quad}p{5cm}@{}c@{\quad\ \ \ }l@{\quad}p{4.8cm}@{}c@{}}
    \toprule
    & \multicolumn{2}{c}{\textbf{Original}} & \multicolumn{3}{c}{\textbf{Reformulation}} & \multicolumn{3}{c}{\textbf{Semantic Change}} \\
    \cmidrule(r{.6cm}){2-3} \cmidrule(r{.6cm}){4-6} \cmidrule(l{-.05cm}r{-.05cm}){7-9}
     & \multirow{1}{*}[-.5em]{\textbf{Name}} & \multirow{1}{*}[-.5em]{\textbf{Example}} & \multirow{1}{*}[-.5em]{\textbf{Name}} & \multirow{1}{*}[-.5em]{\textbf{Example}} & \textbf{Answer} & \multirow{1}{*}[-.5em]{\textbf{Name}} & \multirow{1}{*}[-.5em]{\textbf{Example}} & \textbf{Answer}
    \\
    & & & & & \textbf{Changes} & & & \textbf{Changes}
    \\
    \midrule
    \midrule
    \multirow{3}{*}[-.25em]{\rotatebox[origin=c]{90}{\textbf{Language}}} & \multirow{3}{*}[-.5em]{\textbf{O}} & \multirow{3}{*}[-.5em]{Which animal has longer fur?} & \textbf{LR-I} & State which animal has longer fur. & \bxmark & \textbf{LS-N} & Which animal has not longer fur? & \bcmark \\
    & & & \textbf{LR-C} & Which fur is longer? & \bxmark & \textbf{LS-A} & Which animal has shorter fur? & \bcmark \\
    & & & \multirow{3}{*}[+.5em]{\textbf{LR-V}} & In the presented picture, which animal has longer fur? & \multirow{3}{*}[+.5em]{\bxmark} & \multirow{3}{*}[+.5em]{\textbf{LS-M}} & \multirow{3}{*}[+.5em]{Which animal has less long fur?} & \multirow{3}{*}[+.5em]{\bcmark} \\
    \midrule
    \multirow{3}{*}[-.4cm]{\rotatebox[origin=c]{90}{\textbf{Vision}}} & \multirow{3}{*}[-1.em]{\textbf{O}} & \multirow{3}{*}[-.4cm]{\includegraphics[height=1cm]{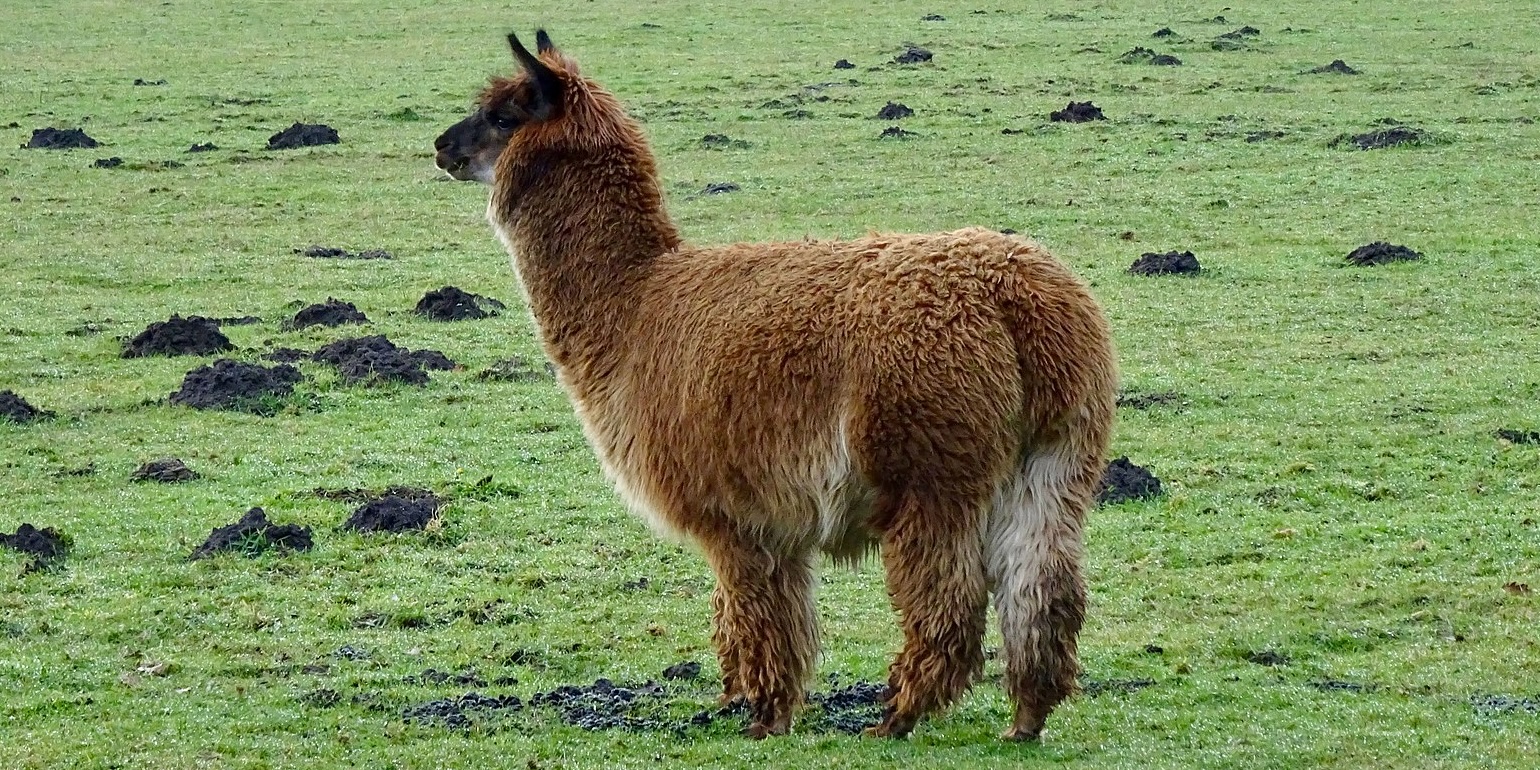}\includegraphics[height=1cm]{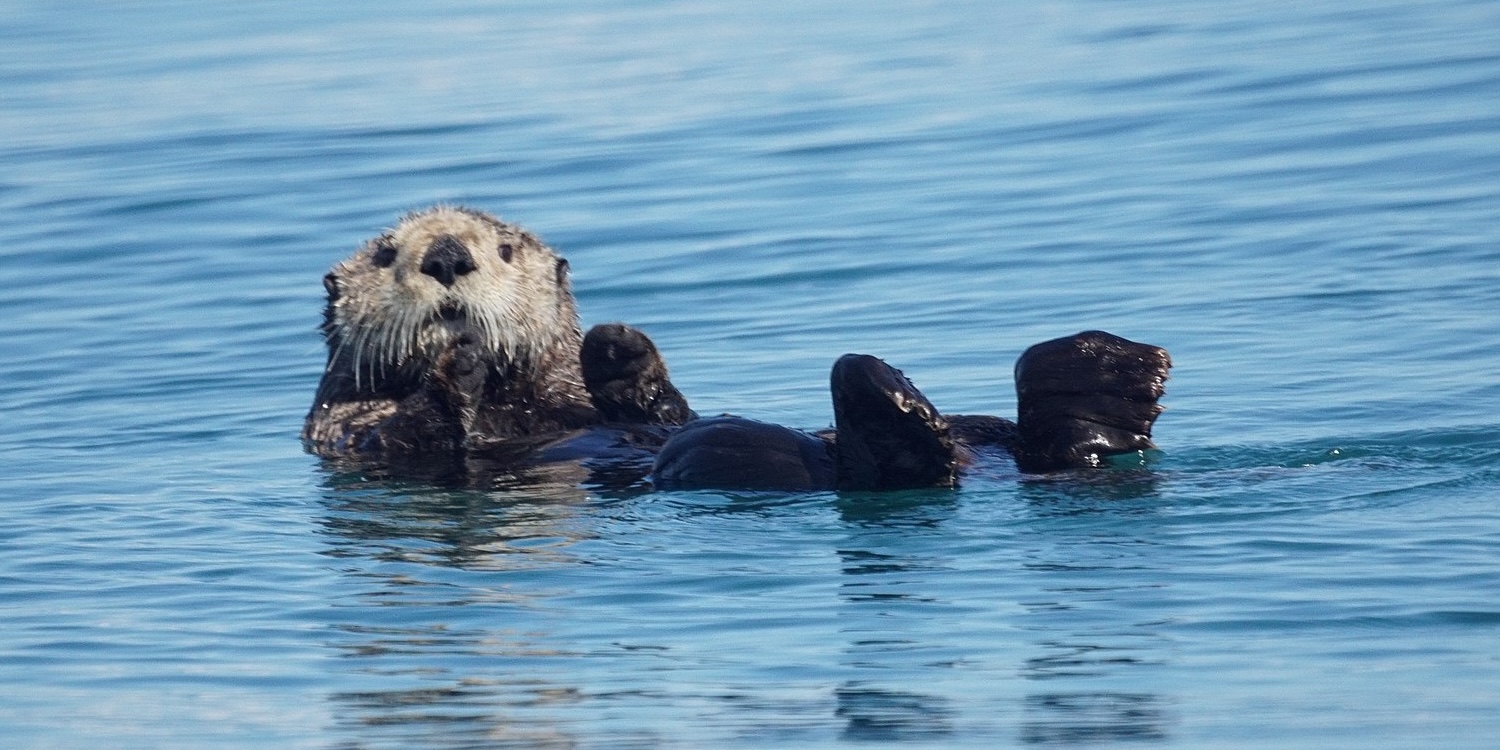}} & \multirow{3}{*}[.75cm]{\textbf{VR-B}} & \includegraphics[height=1cm]{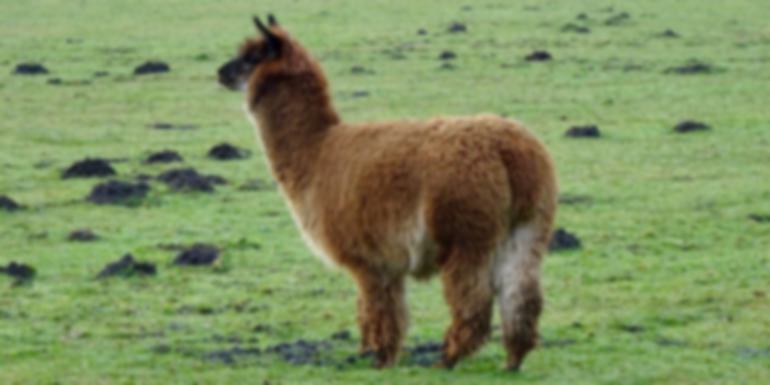}\includegraphics[height=1cm]{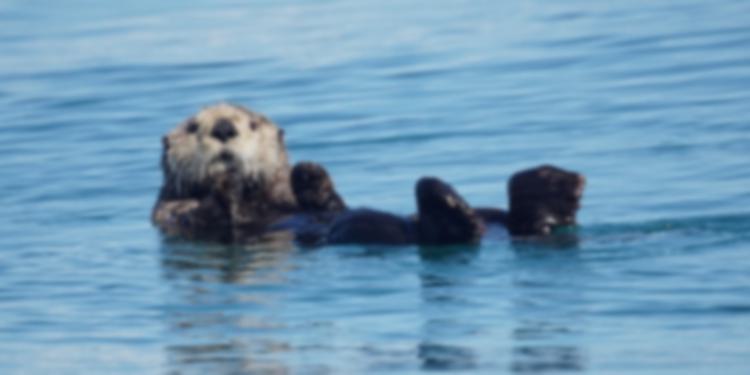} & \multirow{3}{*}[.75cm]{\bxmark} & \multirow{3}{*}[.75cm]{\textbf{VS-S}} & \includegraphics[height=1cm]{images/hair_short02.jpg}\includegraphics[height=1cm]{images/hair_long04.jpeg} & \multirow{3}{*}[.75cm]{\bcmark} \\
    & & & \multirow{3}{*}[.75cm]{\textbf{VR-L}} & \includegraphics[height=1cm]{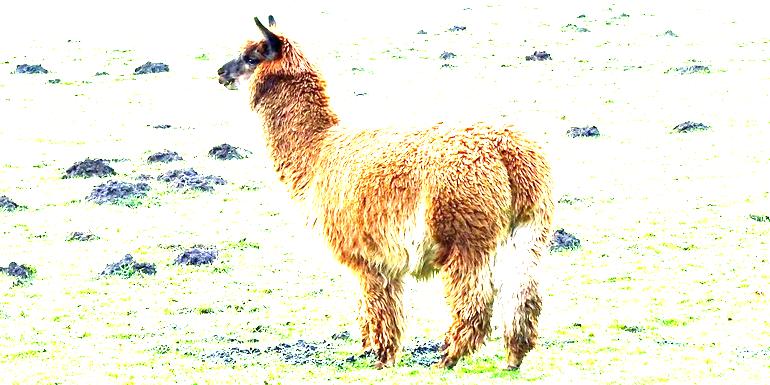}\includegraphics[height=1cm]{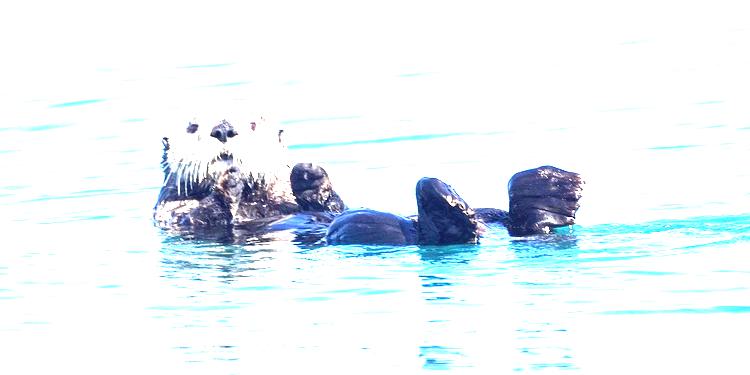} & \multirow{3}{*}[.75cm]{\bxmark} & \multirow{3}{*}[.75cm]{\textbf{VS-E}} & \includegraphics[height=1cm]{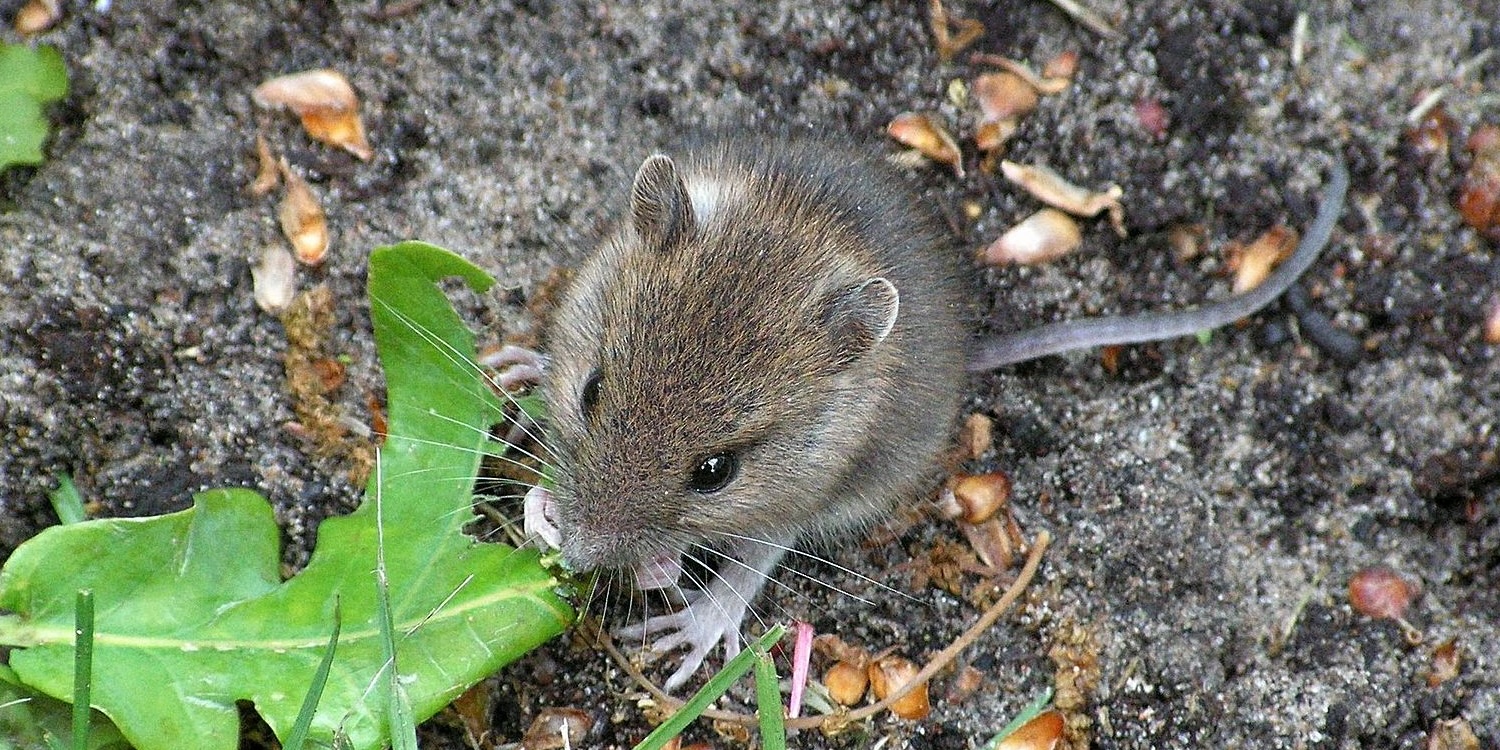}\includegraphics[height=1cm]{images/hair_short02.jpg} & \multirow{3}{*}[.75cm]{\bcmark} \\
    & & & \multirow{3}{*}[.75cm]{\textbf{VR-R}} & \hspace*{1.5cm}\includegraphics[angle=90,height=1cm]{images/hair_long04.jpeg}\includegraphics[angle=90,height=1cm]{images/hair_short02.jpg} & \multirow{3}{*}[.75cm]{\bxmark} & & & \\
    \bottomrule
    \end{tabular}
    }
    \vspace{-.2cm}
    \caption{\textcolor{promptcol}{\textbf{Prompt variations in \framework{}}}. Our VLM prompt variations alter language and vision through reformulations and semantic changes on comparative-style question~\cite{kil2024compbench}. Original question: \enquote{Which animal has longer fur? (A)~Left (B)~Right} with images shown above. \emph{Answer Change} implies the expected answer changed compared to the original question. \emph{Prompt variation types:} LR - Language Reformulation, LS - Language Semantic, VR - Vision Reformulation, VS - Vision Semantic. \emph{Variations:} I - Instruction, C - Concise, V - Verbose, N - Negation, M - MoreLess, A - Antonyms, B - Blur, L - Lighting, R - Rotate, S - Swap, E - Exchange.}
    \label{tab:prompt-var-overview}
    \vspace{-.5\baselineskip}
\end{table*}

We established three ingredients for meaningful VLM prompt sensitivity analyses:
Understandable prompt variations, a model reliability score and score comparability.
Below, we describe how we design them within \framework{}.

Because prompt variations may alter the correct answer, we focus on Multiple-Choice Visual Question Answer tasks (MC-VQA)~\cite{liu2023mmbench,kil2024compbench,fu2024isobench} because it allows us to track the correct answers within its discrete options.
For example, for the question \enquote{How many otters are there?} and an image of otters, among the potential answers, \enquote{(A)~3 (B)~6 (C)~1}, we know that only one is correct. 
Contrastively, generative tasks have non-discrete answers, which do not allow direct answer adaptations and require an extra LLM for scoring.

\subsection{Generating Prompt Variations}
\label{sec:promptvariations}

To understand VLM behavior for different user prompts, we introduce a comprehensive set of 11 realistic variations for the language and vision components of VLM prompts.
In \framework{}, we establish two types of practical variations: \emph{Prompt reformulations} that keep the originally correct answer, and \emph{semantic changes} that change the correct answer to test if the prompt meaning was understood.
Both types apply to either the prompt's language or vision component.
Below we describe the generation of all four prompt variation types, examples in \cref{tab:prompt-var-overview}.
As reformulations are answer-preserving text~\cite{schiappa2022robustness} and image~\cite{hendrycks2018benchmarking} perturbations,
\vspace*{-1.35\baselineskip}
\begin{align*}
&
\small
\text{reformulation }^\text{vision}_\text{language}  \subset \text{perturbation }^\text{image}_\text{text}  \subset \text{prompt variation} ,
\end{align*}

\vspace*{-.35\baselineskip}
\noindent
we select representative concepts from~\cite{schiappa2022robustness,hendrycks2018benchmarking} that maximize prompt variability while being human-plausible.
For semantic changes, introduce tailord variations with \framework{}.

\para{Language reformulation}
We consider three reformulations of the original question that do \emph{not} change the answer: 
(1) Question to \underline{I}nstructions (LR-I);
(2) \underline{C}oncise questions (LR-C) with less words;
and (3) \underline{V}erbose questions (LR-V) with more words.
We use LLaMA3-70B~\cite{lm3} to reformulate the original text prompts (see LLaMA prompts in Supp).

\para{Language semantics}
Here, we focus on negations because negating the question changes the expected answer, and for most questions the correct answers after negation are those that were originally wrong.
Specifically, we consider
(1) negation by adding a \underline{N}ot (LS-N), which works for most questions, and
(2) substituting \underline{A}ntonyms (LS-A) for question attributes or states.
If question phrasing allows, we also
(3) swap the word \enquote{\underline{M}ore} with \enquote{less} (LS-M).
Again, we rephrase with LLaMA3-70B and manually check results. Also, we thoroughly clean MMBench to enable automated language prompt variations.

Semantic changes produce multiple correct answers for original prompts with more than two answers.
During VLM evaluation, we accept responses if they contain one or several of these correct answers.
We directly count a response as wrong if it contains the originally correct answer.

\para{Vision reformulation}
As reformulations on the image, we consider (1) \underline{B}lurring (VR-B); (2) \underline{L}ighting changes (VR-L) through brightening; and (3) \underline{R}otations by 90$^\circ$ (VR-R).
These image modifications maintain the expected answer, because they retain the order of attributes in comparisons.

\para{Vision semantics}
Changing the correct answer through purely visual modifications is only possible for structured image-question pairs, in which we ask comparative questions about \textit{two} collocated images.
In this setting, the answer changes if (1) the images are \underline{S}wapped (VS-S), or if (2) the correct image is \underline{E}xchanged (VS-E) for one that answers the question incorrectly. 
See \cref{tab:prompt-var-overview} for examples. For VS-E, we manually select exchangeable images from the datasets, effectively creating new datasets. Details in Supp.

\subsection{Reliability Score}

Given our varied set of prompt variations, we need expressive scores to quantify VLM prompt sensitivity.
Within \framework{}, we use the existing metrics, accuracy, certainty and consistency, and propose a new reliability score.
\framework{}'s reliability combines accuracy and certainty into a single expressive score, providing two explicit guarantees for both.

\para{Existing scores for prompt sensitivity}
To analyze VLM prompt sensitivity, we distinguish between \textit{per prompt} scores that are calculated for a single prompt $p$, and \textit{prompt variation} scores that are calculated between two prompt variations $p_1$ and $p_2$.
\framework{} includes the per-prompt scores accuracy $\mathit{acc}$ and certainty $\mathit{cert}$, and the variation score consistency $\mathit{cons}$; all defined for a VLM $\mathcal{M}$ that responds to prompt~$p$ with corresponding correct answer(s)~$\mathcal{A}(p)$.

\textbf{\textit{Accuracy}} evaluates if the model answers correctly, $\mathds{1}$ is an indicator function:
$
\mathit{acc}(p) = \mathds{1}[\mathcal{M}(p) \in \mathcal{A}(p)].$

\textbf{\textit{Certainty}} evaluates a model's confidence in its answers.
We calculate it from all possible answers $\mathcal{P}$ and a subset of \emph{likely} predictions $\mathcal{C}$ with the following intuition:
A model is uncertain, $\mathit{cert}\!=\!0$, if $\mathcal{C}$ contains all possible answers and certain, $\mathit{cert}\!=\!1$, if only one answer is likely:
\begin{equation}
\label{equ:cert}
    \mathit{cert}(p) = 1 - \frac{|\mathcal{C}(p)|-1}{|\mathcal{P}(p)|-1}
\end{equation}
To generate $\mathcal{C}$, we use conformal prediction~\cite{angelopoulos2021gentle,kostumov2024uncertainty,ye2024benchmarking} on the softmax scores of possible answers.
Unlike thresholding the scores directly, conformal prediction makes $\mathcal{C}$ contain the correct answer with a certain probability, \eg 90\%.

\textbf{\textit{Consistency}} 
evaluates if model responses agree for two prompt versions $p_1$ and $p_2$~\cite{hudson2019gqa,jang-etal-2022-becel}, independent of their correctness.
For reformulations, the responses should match because the answers are identical.
For semantic changes however, the response should change -- if it was \enquote{B} for $p_1$, it should be anything except \enquote{B} for $p_2$ negated:
\begin{align}
    \mathit{cons}^\text{Reph}(p_1,p_2) &= \mathds{1}[\mathcal{M}(p_1) = \mathcal{M}(p_2)] \\
    \mathit{cons}^\text{Sem}(p_1,p_2) &= \mathds{1}[\mathcal{M}(p_1) \neq \mathcal{M}(p_2)]
\end{align}

\begin{figure}
   \centering
   \includegraphics[width=0.49\columnwidth]{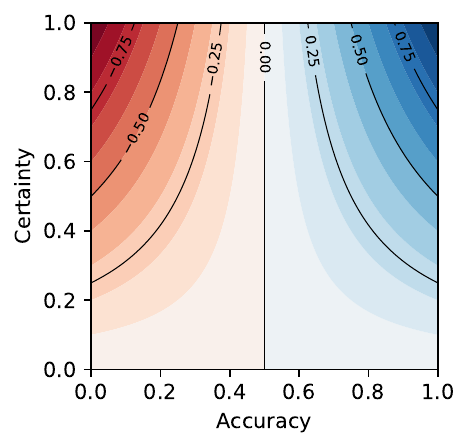}
   \includegraphics[width=0.49\columnwidth]{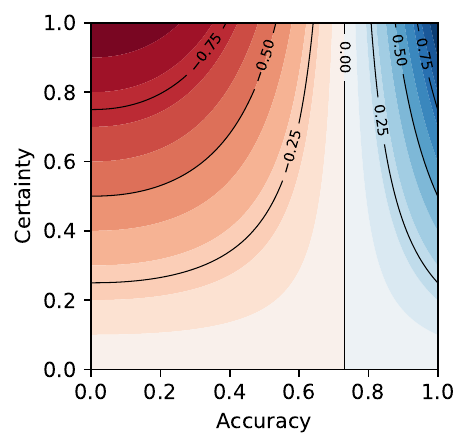}
   \vspace{-.35cm}
   \caption{
   \textcolor{reliabcol}{\textbf{Reliability score in \framework{}.}} The plots visualize how accuracy and certainty are mapped to the reliability score. A reliability of \textcolor{reliabblue}{1 \emph{[blue]}} highlights a confidently correct model, while \textcolor{reliabred}{-1 \emph{[red]}} flags confidently incorrect models. \textit{[Left]} Mapping for a balanced dataset like NYU-Depth V2~\cite{silberman2012NYUDepthv2}, where expected random accuracy is 0.5. \textit{[Right]} Calibrated reliability scores for $\mathit{acc}_\text{rand} = 0.27$, which represents MMBench~\cite{liu2023mmbench}.
   }
    \vspace{-0.35cm}
   \label{fig:Robustness_score}
\end{figure}

\para{Reliability score}
With \framework{}, we introduce a reliability score that acts as a summarization layer on top of an existing accuracy and certainty score.
Intuitively, a reliability of 1 should highlight models that are accurate and confident (highly reliable), while –1 should flag confident but inaccurate (highly unreliable) models. We achieve this by defining
\begin{equation}
    \label{equ:reliab}
    \mathit{rel} = (2 \cdot \mathit{acc} -1) \cdot \mathit{cert}.
\end{equation}
This score is 0 for uncertain models with $\mathit{cert}\!=\!0$, or models with accuracy 0.5.
\cref{fig:Robustness_score} illustrates the mapping from accuracy and certainty to reliability.
The main benefit of this reliability score is its intuitive combination of accuracy and certainty into a single number that contains the most important insights from both metrics.
Furthermore, it comes with two explicit guarantees for accuracy and certainty in~\cref{equ:guarant}, which we introduce after calibration in the next section.

\subsection{Calibration of Metrics}
\label{sec:calibmetrics}

To evaluate the performance of VLMs across datasets and prompt variations in \framework{}, we would like to directly compare the performance scores -- accuracy, certainty, consistency and reliability.
In practice, however, this is problematic as datasets and prompt variations may come with different difficulties in the form of expected random performance.
Across datasets, fewer answer options increase the chances of picking the correct one.
Across prompt variations, negating the question often creates more correct answers, which also increases the chance of guessing a correct one.
To compare the performance across datasets and prompts, we therefore need to correct for these differences in expected random performance.
Hence, in \framework{} we introduce a calibration mechanism for improved comparability of accuracy, consistency, certainty and reliability scores across datasets and prompts with varying difficulty.

\begin{figure}
   \centering
   \begin{tabular}{@{}c@{}c@{}c@{}}
    & \hspace{1.5em}\footnotesize Accuracy \textbf{Before Calibration} & \hspace{1em}\footnotesize Accuracy \textbf{After Calibration}
    \\
    \multirow{1}{*}[1.0cm]{\rotatebox[origin=c]{90}{\footnotesize MMBench}} 
    & \multicolumn{2}{c}{\hspace*{.01\linewidth}\includegraphics[width=.93\linewidth]{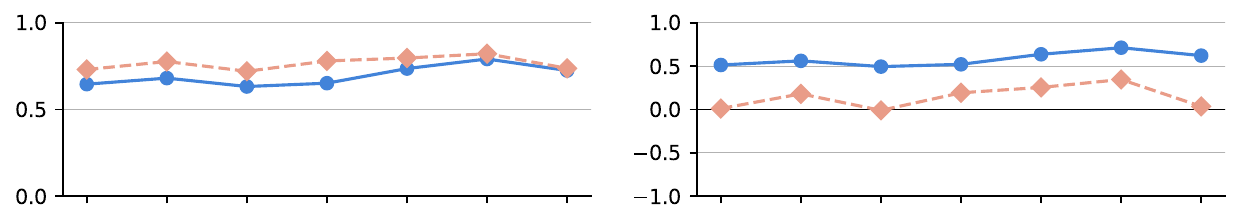}}
    \\
    \multirow{1}{*}[1.75cm]{\rotatebox[origin=c]{90}{\footnotesize NYU-Depth}}
    & \multicolumn{2}{c}{\includegraphics[width=.94\linewidth]{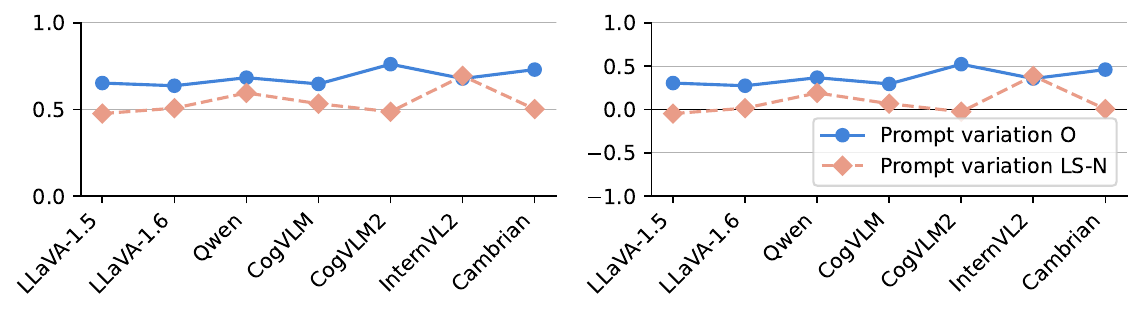}}
    \\
   \end{tabular}
   \vspace{-0.3cm}
   \caption{
   \textcolor{calibcol}{\textbf{Calibration effect in \framework{}.}} On MMBench~\cite{liu2023mmbench} before calibration, models appear more accurate on negated prompts LS-N \textcolor{plotorange}{\textit{[Orange, dashed]}} than on the original prompts O \textcolor{plotblue}{\textit{[Blue, solid]}}, LS-N$>$O.
   After calibrating by measuring the improvement over the expected random performance, the order switches to O$>$LS-N.
   Why should we expect VLMs to be better on O than LS-N?
   We take a look at NYU-Depth~V2~\cite{silberman2012NYUDepthv2} -- a balanced dataset with two potential answers, where VLMs have the same expected random performance on O and LS-N.
   Here, VLMs still perform worse on negations than on the original prompt, showing that O$>$LS-N is the expected trend.
   Because calibration uses the same random performance of 0.5 for O and LS-N, the accuracy-order O$>$LS-N is maintained on NYU-Depth after calibration.
   In contrast, MMBench has 3 answers, giving random selection a lower chance to find the one correct answer for O than to find one of two for LS-N.
   Because the random performance for O and LS-N is different on MMBench, the order changes after calibration.
   }
    \vspace{-0.3cm}
   \label{fig:calib_MMBench_CompBench}
\end{figure}

\para{Generic calibration -- Accuracy and consistency}
Our calibrated scores $s_\text{calib}$ measure the improvement over the expected random performance $s_\text{rand}$ of the pure score $s$:
\begin{equation}
    s_\text{calib} = \begin{cases}
    \frac{s-s_\text{rand}}{1-s_\text{rand}} \quad \text{for } s \geq s_\text{rand},\\
    \frac{s-s_\text{rand}}{s_\text{rand}}   \quad \text{for } s < s_\text{rand},
    \end{cases}
    s_\text{calib} \in [-1,1].
\end{equation}
Calibrated scores are $1$ for ideal, $0$ for random and $-1$ for worst-case performance.
A model with $s_\text{calib}\!=\!0.6$ closes 60\% of the gap between random and ideal performance.
This procedure is directly applicable to accuracy and consistency, once the expected random performance is known.
In \cref{fig:calib_MMBench_CompBench} we illustrate the effect of calibration.
The results also validate our calibration procedure, because we observe the trends from the balanced NYU-Depth V2 data~\cite{silberman2012NYUDepthv2} on the imbalanced MMBench data~\cite{liu2023mmbench} \textit{only after calibration}.

\para{Calibration for certainty via conformal prediction}
To calculate the certainty in \cref{equ:cert}, we need the conformal prediction set $\mathcal{C}$ per prompt.
We follow \cite{kostumov2024uncertainty} to obtain prediction sets with an error rate of less than 10\% and use the LAC scoring function \cite{sadinle2019lac}.
This generates reliable certainties for prompts with a single correct answer -- but not for semantically changed prompts with multiple correct answers, where the model is correct if it predicts any of them.
We solve this by treating all correct answers as one, \ie sum their softmax scores, and then applying conformal prediction as described above.

\para{Calibration for reliability}
To make the reliability comparable across datasets with varying difficulties, we shift the neutral reliability score of 0 to the measured random accuracy $\mathit{acc}_\text{rand}$.
Thus, the reliability is positive if the model is more accurate than random, and otherwise negative.
For calibration, $\mathit{acc}$ is replaced by $\mathit{acc}^m$ with $m = \frac{\log(2)}{\log(1/\mathit{acc}_\text{rand})}$ in \cref{equ:reliab}, \cf~\cref{supp.sec:reliability score}.
\cref{fig:Robustness_score} visualizes the effect.

\para{Reliability interpretation and guarantees}
With this intuitive understanding of calibration, \framework{}'s reliability definition brings two explicit accuracy and certainty guarantees that can be directly read from a given score.
Their phrasing slightly varies for positive and negative scores:
\begin{equation}
    \label{equ:guarant}
    \mathit{cert} \geq |\mathit{rel}|, \quad  \ 
    \mathit{acc}_\text{calib}
    \begin{cases}
    \geq \mathit{rel} & \quad \text{for } \mathit{rel} > 0 \\
    \leq \mathit{rel} & \quad \text{for } \mathit{rel} < 0
    \end{cases}.
\end{equation}
A \textit{positive} reliability of 0.3 guarantees at least 30\% improvement over $\mathit{acc}_\text{rand}$ with at least 30\% certainty.
For \textit{negative} scores, the guaranteed performance improvement becomes a guaranteed reduction:
Reliability –0.5 guarantees at least 50\% performance reduction over $\mathit{acc}_\text{rand}$ with at least 50\% certainty.
These two guarantees allow at-a-glance comparison of \framework{}'s robustness score across different datasets and prompt variations.

\begin{table*}[t]
\centering
\scalebox{0.66}{
\begin{tabular}{lll@{\quad\quad\quad}rrrrrrrr@{\quad\quad\quad}r@{\quad\quad\quad}r@{\quad\quad\quad}rrrrrrrr}
\toprule
& & & \multicolumn{8}{c}{\textbf{Reliability Calibrated}} & \textbf{Acc C} & \textbf{Cert C} & \multicolumn{8}{c}{\textbf{Consistency Calibrated}} \\
\cmidrule(r{1cm}){4-11} \cmidrule(r{1cm}){12-12} \cmidrule(r{1cm}){13-13} \cmidrule(r){14-21}
\multicolumn{3}{l}{\textbf{Prompt Variation}} & M-S & M-A & V-S & V-A & NYU & Fas & MMB & \textbf{AVG} & \textbf{AVG} & \textbf{AVG} & M-S & M-A & V-S & V-A & NYU & Fas & MMB & \textbf{AVG} \\
\midrule
\midrule
\multicolumn{2}{l}{\textbf{Original}} &
\textbf{O}     & 0.52 & 0.12 & 0.24 & 0.23 & 0.13 & 0.32 &  0.48 & 0.29     & 0.49 & 0.54 &      \\
\midrule
\multirow{6}{*}[-.25em]{\rotatebox[origin=c]{90}{\textbf{Language}}}
& \multirow{3}{*}[0em]{\rotatebox[origin=c]{90}{\textbf{Ref.}}} &
\textbf{LR-I}  & 0.51 & \textbf{0.07} & 0.22 & 0.16 & 0.13 & 0.29 &  0.47 & 0.26     & 0.46 & 0.50 &     0.83 & \textbf{0.68} & 0.74 & 0.73 & 0.79 & 0.80 & 0.87 & 0.78 \\
& &
\textbf{LR-C}  & 0.41 & 0.13 & 0.17 & 0.15 & 0.11 & 0.28 &  0.45 & 0.24     & 0.45 & 0.49 &     0.78 & 0.73 & 0.71 & 0.79 & 0.64 & 0.63 & 0.84 & 0.73 \\
& &
\textbf{LR-V}  & \textbf{0.35} & 0.22 & \textbf{0.14} & \textbf{0.12} & \textbf{0.09} & \textbf{0.09} &  \textbf{0.42} & \textbf{0.21}     & \textbf{0.42} & \textbf{0.43} &     \textbf{0.72} & \textbf{0.69} & \textbf{0.63} & \textbf{0.62} & \textbf{0.49} & \textbf{0.49} & \textbf{0.79} & \textbf{0.63} \\
\cmidrule(r){2-21}
& \multirow{3}{*}[0em]{\rotatebox[origin=c]{90}{\textbf{Sem.}}} &
\textbf{LS-N}  & 0.39 & \textbf{0.06} & 0.22 & 0.18 & 0.04 & 0.25 &  0.05 & 0.17     & 0.39 & 0.37 &     0.58 & 0.50 & 0.42 & 0.44 & 0.06 & 0.37 & 0.06 & 0.35 \\
& &
\textbf{LS-A}  & 0.30 & 0.14 & \textbf{0.11} & \textbf{0.10} & \textbf{0.00} & \textbf{0.05} & \textbf{-0.01} & \textbf{0.10}     & \underline{\textbf{0.24}} & \underline{\textbf{0.30}} &     0.48 & \textbf{0.30} & \textbf{0.22} & \textbf{0.15} & \textbf{-0.34} & \textbf{0.01} & \textbf{-0.16} & \underline{\textbf{0.09}} \\
& &
\textbf{LS-M}  & \textbf{0.15} & 0.09 & 0.21 & 0.14 & 0.01 & 0.17 &   \emph{n.a.} & 0.13     & 0.36 & \underline{\textbf{0.30}} &     \textbf{0.40} & 0.42 & 0.30 & 0.28 & -0.26 & 0.28 & \emph{n.a.} & 0.24 \\
\midrule
\multirow{6}{*}[.5em]{\rotatebox[origin=c]{90}{\textbf{Vision}}}
& \multirow{3}{*}[0em]{\rotatebox[origin=c]{90}{\textbf{Ref.}}} &
\textbf{VR-B}  & 0.41 & \textbf{0.09} & 0.23 & 0.22 & 0.13 & 0.32 &  0.43 & 0.26     & 0.46 & 0.52 &     0.75 & 0.71 & 0.80 & 0.80 & 0.77 & 0.83 & 0.84 & 0.79 \\
& &
\textbf{VR-L}  & \textbf{0.33} & 0.12 & \textbf{0.15} & \textbf{0.20} & 0.09 & 0.19 &  \textbf{0.41} & \textbf{0.21}     & \textbf{0.39} & \textbf{0.45} &     \textbf{0.56} & \textbf{0.63} & \textbf{0.55} & \textbf{0.61} & 0.62 & 0.61 & 0.79 & \textbf{0.63} \\
& &
\textbf{VR-R}  & 0.48 & 0.21 & 0.23 & 0.29 & \textbf{0.08} & \textbf{0.13} &  0.38 & 0.26     & 0.44 & 0.52 &     0.79 & 0.70 & 0.68 & 0.72 & \textbf{0.53} & \textbf{0.56} & \textbf{0.74} & 0.67 \\
\cmidrule(r){2-21}
& \multirow{3}{*}[0.75em]{\rotatebox[origin=c]{90}{\textbf{Sem.}}} &
\textbf{VS-S}  & 0.59 & 0.12 & 0.21 & 0.24 & 0.17 & 0.40 &   \emph{n.a.} & 0.29     & 0.53 & 0.49 &     0.65 & 0.56 & 0.42 & 0.46 & 0.23 & 0.45 & \emph{n.a.} & 0.46 \\
& &
\textbf{VS-E}  & \textbf{0.13} & \textbf{0.19} & \textbf{0.04} & \textbf{0.11} & \textbf{0.16} & \textbf{0.16} &   \emph{n.a.} & \underline{\textbf{0.13}}     & \underline{\textbf{0.33}} & \underline{\textbf{0.36}} &     \textbf{0.13} & \textbf{0.09} & \textbf{-0.04} & \textbf{-0.04} & \textbf{-0.14} & \textbf{-0.02} & \emph{n.a.} &  \underline{\textbf{0.00}} \\
\bottomrule
\end{tabular}
}
\vspace{-0.2cm}
\caption{Most perturbing prompt variations in language and vision, averaged across models. All scores are calibrated, with 1.0 being ideal model performance and 0.0 is random performance. For reliability and consistency we report results on the individual datasets along with averages across datasets, and averages across datasets for accuracy and certainty. See~\cref{tab:supp_promptvar-comp-reliab} for all measures. The consistency is calculated between each varied and the original prompt. Most perturbing prompt per variation class is bold, and across variations underlined. High number indicates model robustness to a prompt variant. Prompt variation acronyms are from~\cref{tab:prompt-var-overview}. MMBench's non-comparative questions about single images neither support more-less reformulation (LS-M), nor image swaps or exchanges (VS-S, VS-E).}
\vspace{-0.15cm}
\label{tab:promptvar-comp-reliab}
\end{table*}

%%%%%%%%% EXPERIMENTS
\section{Experiments}
\label{sec:experiments}

With \framework{} we strive to understand which user prompt variations VLMs are most sensitive to, and if certain models have an outstanding robustness to prompt variations.
Therefore, we evaluate our 11 prompt variations across 7 datasets and 22 models from 7 VLM families.
As model families, we consider LLaVA1.5, LLaVA1.6, Qwen-VL, CogVLM, CogVLM2, InternVL2 and Cambrian~\cite{liu2023improvedllava,liu2024llavanext,bai2023qwen,wang2023cogvlm,hong2024cogvlm2,Chen2024internvl,tong2024cambrian1fullyopenvisioncentric} due to diversity in architecture, input resolutions, LLM sizes and training data.
Our list of 22 VLMs is in~\cref{tab:model-comp-reliab} and the Supp.
As MC-VQA datasets, we include MMBench (MMB)~\cite{liu2023mmbench}, which summarizes multiple datasets, and the comparative datasets MIT-States (M-S), MIT-Attributes (M-A), VAW-States (V-S), VAW-Attributes (V-A), NYU-Depth~V2 (NYU) and  Fashionpedia (Fas)~\cite{isola2015MITStatesAttributes,pham2021VAW,silberman2012NYUDepthv2,jia2020fashionpedia}.
The comparative style was proposed by CompBench~\cite{kil2024compbench} and compares two images and provides two answer options, \enquote{Left} and \enquote{Right}, resulting in balanced datasets with $\mathit{acc}_\text{rand}\!=\!0.5$.
The image comparisons enable our semantic vision variations from~\cref{sec:promptvariations}.
MMBench has one image per prompt -- preventing semantic image variations -- and 3.7 answers on average, leading to $\mathit{acc}_\text{rand}\!=\!0.27$. 
Also, we applied comprehensive prompt cleaning, see Supp., on MMBench to obtain a clean base supporting our framework's text variations. All reported metrics are calibrated.

\subsection {Which Prompts are VLMs Most Sensitive to?}
To answer our first question -- \textit{Which prompt variations are VLMs most sensitive to?} -- we evaluate all proposed prompt variations along their two axes, language and vision; and reformulations and semantic variations.
First, we investigate whether prompt sensitivity exists in VLMs, and if VLMs are more sensitive to vision or language variations.
Second, we explore if reformulations or semantic variations are harder for VLMs, and find interesting parallels across modalities.

\para{Vision vs.\ language prompt variations}
To observe if VLMs are language or vision prompt sensitive, we report the performance of each prompt variation in~\cref{tab:promptvar-comp-reliab}, where each number is an average across all 22 models from~\cref{tab:model-comp-reliab}.
Perturbing prompts cause low reliability, accuracy, certainty and consistency values.
All VLMs are prompt sensitive, and many language and vision prompt variations are destructive.
For example, image exchanges (VS-E) and language negation with antonyms (LS-A) almost reduce the consistency to random guessing $\mathit{cons}_\text{rand}\!=\!0$.
Importantly, we observe a positive correlation between reliability and consistency: models that are unreliable on prompts are also inconsistent to the same prompts. 
% Prompts that lead to unreliable models also cause inconsistencies to the original prompt.
Among language variations, rephrased verbose prompts (LR-V) and antonyms (LS-A) are most confusing to models across metrics.
Among vision variations, rephrasing through lighting changes (VR-L) and image exchanges (VS-E) are most destructive.
Interestingly, for image swaps (VS-S), the average model accuracy and reliability improve, indicating model confusion towards spatial arrangements. 
% an answer-bias towards either \enquote{Left} or \enquote{Right}.
Across language and vision, we find that simple rephrasing like instructions instead of questions are tolerated more than prompt variations that change the expected answer.
Overall, our evaluation demonstrates that VLMs are prompt sensitive to language and vision variations, and similarly vulnerable in both modalities.

\begin{figure}
   \centering
   \includegraphics[width=\linewidth]{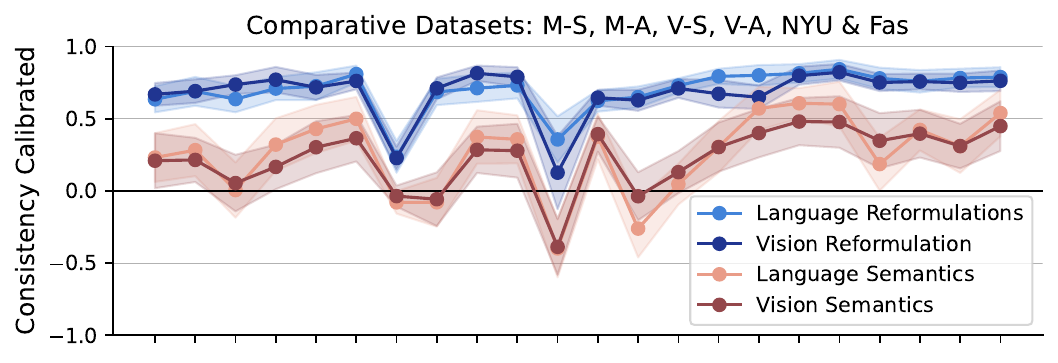}
   \includegraphics[width=\linewidth]{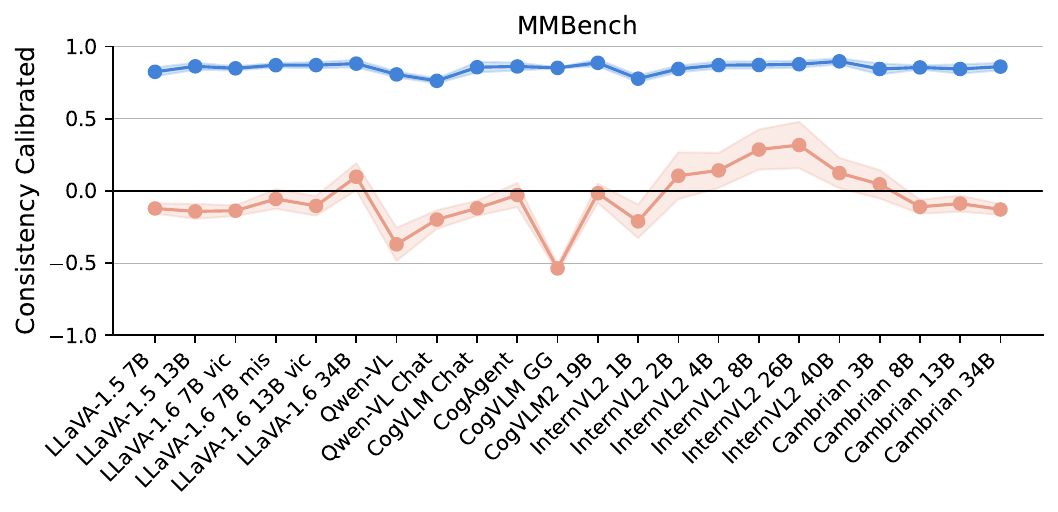}
   \vspace{-0.6cm}
   \caption{
   VLMs understand reformulations \textcolor{plotblue}{\emph{[blue]}} better than semantic changes \textcolor{plotorange}{\emph{[orange]}}, across language and vision modalities. Trends are aligned on the six comparative datasets and MMBench.}
   \vspace{-0.25cm}
   \label{fig:comp_semvsref}
\end{figure}

\begin{table*}[t]
\begin{minipage}[]{0.64\linewidth}\centering
\scalebox{0.5}{
\begin{tabular}{l@{\quad}r@{\ \ }r@{\ \ }r@{\ \ }r@{\ \ }r@{\ \ }r@{\ \ }r@{\quad}r@{\quad\quad}r@{\quad\quad}r@{\quad\quad}r@{\ \ }r@{\ \ }r@{\ \ }r@{\ \ }r@{\ \ }r@{\ \ }r@{\quad}r@{}}
\toprule
& \multicolumn{8}{c}{\textbf{Reliability Calibrated}} & \textbf{Acc C} & \textbf{Cert C} & \multicolumn{8}{c}{\textbf{Consistency Calibrated}} \\
\cmidrule(r{.5cm}){2-9} \cmidrule(r{.5cm}){10-10} \cmidrule(r{.5cm}){11-11} \cmidrule(r{.1cm}){12-19}
\textbf{Model} & M-S & M-A & V-S & V-A & NYU & Fas & MMB & \textbf{AVG} & \textbf{AVG} & \textbf{AVG} & M-S & M-A & V-S & V-A & NYU & Fas & MMB & \textbf{AVG} \\
\midrule
\midrule
LLaVA-1.5 7B      & 0.33 & 0.09 & 0.12 & 0.12 & 0.06 & 0.12 & 0.28 & 0.16        & 0.35 & 0.38 &         0.62 & 0.40 & 0.48 & 0.52 & 0.29 & 0.42 & 0.60 & 0.48 \\
LLaVA-1.5 13B     & 0.27 & 0.17 & 0.17 & 0.14 & 0.08 & 0.11 & 0.31 & 0.18        & 0.38 & 0.41 &         0.62 & 0.56 & 0.52 & 0.53 & 0.23 & 0.50 & 0.60 & 0.51 \\
\midrule
LLaVA-1.6 7B vic  & 0.34 & 0.09 & 0.15 & 0.16 & 0.12 & 0.11 & 0.30 & 0.18        & 0.37 & 0.41 &         0.54 & 0.34 & 0.41 & 0.53 & 0.36 & 0.14 & 0.60 & 0.42 \\
LLaVA-1.6 7B mis  & 0.36 & 0.08 & 0.20 & 0.28 & 0.06 & 0.19 & 0.35 & 0.22        & 0.43 & 0.44 &         0.56 & 0.56 & 0.48 & 0.62 & 0.31 & 0.59 & 0.62 & 0.54 \\
LLaVA-1.6 13B vic & 0.49 & 0.12 & 0.23 & 0.26 & 0.09 & 0.21 & 0.34 & 0.25        & 0.47 & 0.46 &         0.76 & 0.57 & 0.57 & 0.65 & 0.30 & 0.54 & 0.62 & 0.57 \\
LLaVA-1.6 34B     & 0.53 & 0.25 & 0.32 & \textbf{0.36} & 0.15 & 0.25 & 0.44 & 0.33        & 0.54 & 0.57 &         0.79 & 0.69 & 0.60 & 0.68 & 0.41 & 0.61 & 0.62 & 0.63 \\
\midrule
Qwen-VL           & 0.03 & 0.01 & 0.03 & 0.06 & 0.04 & 0.04 & 0.21 & 0.06        & 0.15 & 0.29 &         0.00 & 0.14 & 0.11 & 0.24 & 0.16 & -0.06 & 0.42 & 0.14 \\
Qwen-VL Chat      & 0.11 & 0.05 & 0.09 & 0.15 & 0.04 & 0.05 & 0.25 & 0.11        & 0.27 & 0.32 &         0.31 & 0.40 & 0.40 & 0.42 & 0.35 & 0.21 & 0.52 & 0.37 \\
\midrule
CogVLM Chat       & 0.48 & 0.12 & 0.32 & 0.22 & 0.09 & 0.34 & 0.26 & 0.26        & 0.49 & 0.47 &         0.71 & 0.63 & 0.67 & 0.60 & 0.30 & 0.50 & 0.62 & 0.58 \\
CogAgent          & 0.42 & 0.12 & 0.21 & 0.14 & 0.10 & 0.32 & 0.23 & 0.22        & 0.46 & 0.44 &         0.70 & 0.60 & 0.59 & 0.56 & 0.40 & 0.52 & 0.64 & 0.57 \\
CogVLM GG         & -0.10 & -0.07 & -0.26 & -0.23 & -0.04 & -0.15 & 0.02 & -0.12       & -0.15 & 0.40 &         0.06 & -0.01 & -0.06 & -0.12 & -0.10 & -0.06 & 0.45 & 0.02 \\
\midrule
CogVLM2 19B       & 0.49 & 0.11 & 0.20 & 0.20 & 0.16 & 0.42 & 0.26 & 0.26        & 0.49 & 0.49 &         0.68 & 0.68 & 0.51 & 0.38 & 0.30 & 0.58 & 0.39 & 0.50 \\
\midrule
InternVL2 1B      & 0.13 & 0.03 & 0.01 & 0.04 & 0.05 & 0.08 & 0.30 & 0.09        & 0.20 & 0.28 &         0.32 & 0.40 & 0.31 & 0.18 & 0.15 & 0.26 & 0.54 & 0.31 \\
InternVL2 2B      & 0.23 & 0.15 & 0.11 & 0.11 & 0.09 & 0.14 & 0.37 & 0.17        & 0.38 & 0.38 &         0.48 & 0.53 & 0.44 & 0.42 & 0.29 & 0.40 & 0.64 & 0.46 \\
InternVL2 4B      & 0.35 & 0.21 & 0.23 & 0.18 & 0.13 & 0.29 & 0.41 & 0.26        & 0.47 & 0.49 &         0.67 & 0.61 & 0.50 & 0.59 & 0.34 & 0.52 & 0.64 & 0.55 \\
InternVL2 8B      & 0.51 & \textbf{0.23} & 0.30 & 0.24 & 0.11 & 0.34 & 0.48 & 0.32        & 0.56 & 0.52 &         0.73 & 0.72 & 0.61 & 0.61 & 0.37 & 0.70 & 0.70 & 0.63 \\
InternVL2 26B     & 0.67 & 0.21 & 0.31 & 0.30 & 0.16 & \textbf{0.50} & 0.52 & 0.38        & 0.61 & \textbf{0.58} &         0.83 & 0.77 & 0.69 & 0.70 & \textbf{0.47} & 0.69 & \textbf{0.72} & 0.70 \\
InternVL2 40B     & \textbf{0.69} & 0.20 & \textbf{0.33} & 0.33 & \textbf{0.17} & 0.48 & \textbf{0.57} & \textbf{0.40}        & \textbf{0.65} & 0.57 &         \textbf{0.86} & \textbf{0.81} & \textbf{0.74} & 0.67 & 0.43 & \textbf{0.72} & \textbf{0.72} & \textbf{0.71} \\
\midrule
Cambrian 3B       & 0.48 & 0.16 & 0.18 & 0.11 & 0.10 & 0.23 & 0.39 & 0.24        & 0.45 & 0.45 &         0.72 & 0.55 & 0.57 & 0.50 & 0.28 & 0.57 & 0.62 & 0.54 \\
Cambrian 8B       & 0.52 & 0.16 & 0.24 & 0.28 & 0.12 & 0.26 & 0.40 & 0.28        & 0.51 & 0.49 &         0.80 & 0.69 & 0.54 & 0.57 & 0.43 & 0.57 & 0.60 & 0.60 \\
Cambrian 13B      & 0.51 & 0.20 & 0.23 & 0.20 & 0.11 & 0.25 & 0.38 & 0.27        & 0.49 & 0.49 &         0.76 & 0.62 & 0.62 & 0.53 & 0.29 & 0.53 & 0.61 & 0.57 \\
Cambrian 34B      & 0.55 & 0.18 & 0.28 & 0.25 & 0.13 & 0.27 & 0.46 & 0.30        & 0.52 & 0.52 &         0.84 & 0.77 & 0.59 & \textbf{0.71} & 0.41 & 0.59 & 0.60 & 0.64 \\
\bottomrule
\end{tabular}
}
\end{minipage}
\hfill
\begin{minipage}[]{0.36\linewidth}\centering
\includegraphics[width=1\linewidth]{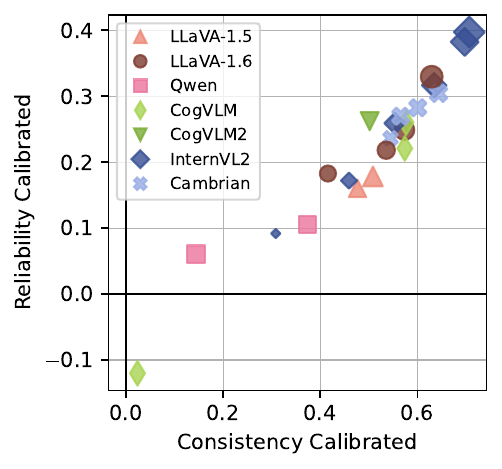}
\end{minipage}
\vspace{-.2cm}
\caption{Most robust VLMs. \emph{[Left]} Reliability, accuracy, certainty and consistency measured across models and datasets, and averaged over prompt variants, also see~\cref{tab:supp-model-comp-reliab}. High numbers indicate high robustness. \emph{[Right]} Reliability versus consistency across all datasets, model size is indicated by marker size. The most robust model family to prompt changes is InternVL2, with 40B as best, also see~\cref{fig:supp_model-winner_fourvariations}.}
\vspace{-0.15cm}
\label{tab:model-comp-reliab}
\end{table*}

\para{Reformulation vs.\ semantic prompt variations}
Motivated by the observation from~\cref{tab:promptvar-comp-reliab} that reformulated prompts are more tolerated by VLMs than semantic variations, we further analyze these variation families. 
In~\cref{fig:comp_semvsref}, we show the consistencies for individual VLMs, averaged across prompt pairs that only modify (1) language reformulations (LR), (2) language semantics (LS), (3) vision reformulations (VR), and (4) vision semantics (VS). 
We analyze all four variations across the six comparative datasets, and only report language variations on MMBench because it does not support vision semantic changes.
On the comparative datasets, we observe a significant gap between reformulations \textcolor{plotblue}{\emph{[blue]}} and semantics changes \textcolor{plotorange}{\emph{[orange]}}.
Also, the behavior of language and vision reformulations is almost perfectly aligned -- and similarly for semantic changes.
This is remarkable because our prompt reformulations and semantic changes are not designed to be identical across vision and language -- the only shared characteristic across reformulations and semantic changes is whether they change the expected answer or not.
Noteably, observations on the comparative datasets are mirrored in MMBench: from the consistency gap between reformulations and semantics to the poor performances of Qwen-VL and CogVLM-GG.
Conclusively, our prompt variation analysis with \framework{} shows that VLMs inherit LLM prompt sensitivity -- and do not consistently understand the semantic prompt meaning.

\subsection {Which VLM is Most Prompt Agnostic?}

We have seen that both language and vision prompt variations lead to volatile behavior in VLMs.
Therefore, we now approach our second question -- \textit{Which VLM class is most agnostic to prompt variations?} -- by focusing on the per-model performance across benchmarks and prompt variations.
Here, we initially evaluate all models on all datasets using reliability and calibrated consistency.
Then, we explore trends in VLM families to analyze which aspect of VLMs is more related to prompt sensitivity.

\para{VLM consistency and reliability across datasets}
For all 22 models, we summarize the prompt sensitivity results from \framework{} in~\cref{tab:model-comp-reliab}.
We report calibrated reliability, accuracy, certainty and consistency, where large values indicate good models, averaged over all datasets, with per-dataset results for reliability and consistency.
Due to our calibrated metrics, we can directly compare the averaged consistency and reliability scores across datasets to find the most consistent and reliable models.
Generally, models are less reliable for Attribute questions (M-A, V-A) than for States (M-S, V-S), indicating a better general understanding of objects than details.
Among models, InternVL2-40B performs generally best, while Qwen models and CogVLM GG perform worst.
\emph{[Right]} of~\cref{tab:model-comp-reliab} we plot a common ranking of models by their reliability and consistency, and color-code the model family.
Again as in~\cref{tab:promptvar-comp-reliab}, reliability and consistency are well correlated, 
showing no inherent trade-off when selecting the most reliable vs consistent model.
Interestingly, model family seems to be a better indicator of reliability and consistency than model size: 
For example, InternVL2-2B is on par with LLaVA-1.5 13B w.r.t.\ average reliability and consistency, and models within the same family like LLaVA-1.5 or Cambrian cluster in the plot.
However, within the same family, larger models perform better. 
Overall, in the InternVL2 family its 40B model stands out for exceptional reliability and consistency for varying prompts.

\begin{figure}
   \centering
   \includegraphics[height=0.385\linewidth]{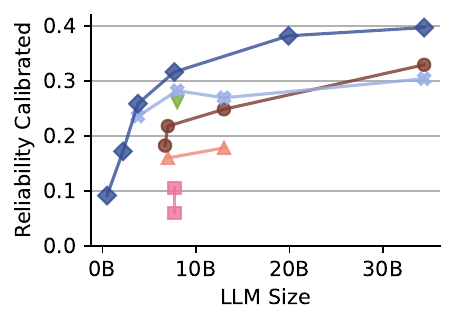}
   \includegraphics[height=0.385\linewidth]{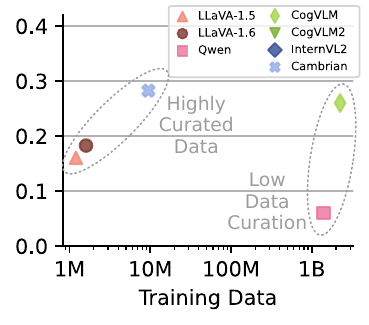}
   \vspace{-.3cm}
    \caption{\emph{[Left]} Larger models within the same family are less prompt sensitive. \emph{[Right]} More high-quality data yields more prompt-agnostic models. Comparison shows models with 7B or 8B LLMs per family; with Qwen-VL for Qwen, LLaVA-1.6 7B vic for LLava-1.6 and CogVLM Chat for CogVLM. Note the log-scale, also see~\cref{fig:supp_model-size}.}
\label{fig:model-size}
\vspace*{-0.5cm}
\end{figure}

\para{Causes for sensitivity differences in VLMs}
Motivated by the large performance differences among VLM families, we explore four potential reasons for increased resilience to prompt variations: LLM size, VLM size, image resolution and training data.
Because some information are unreleased, we cannot collect all of this data for all tested VLMs, limiting our analysis to different model selections.
In~\cref{fig:model-size} \emph{[Left]} we show the reliability over LLM size per VLM family.
We see systematic reliability differences between models from different families, and InternVL2 models outperform other models across all sizes.
We observe the same for VLM sizes, see Supp.
This indicates that architecture size does not sufficiently explain prompt sensitivity, because the performance gaps \emph{across} families for the same model size are larger than the gaps between model sizes \emph{within} a family.
Cambrian-8B outperforms its family's performance trend, potentially because of its 2x larger LLM context.
Next, InternVL2 has the largest image resolution of all models.
If image resolutions explain the reliability differences, the reliability differences on language-only variation should be smaller.
However, we observe the same differences when we re-plot~\cref{fig:model-size} \emph{[Left]} for language-only variations in the Supp., which makes image resolution an unlikely explanation.
Regarding training data, the data size corresponds to the reliability order in~\cref{fig:model-size} \emph{[Left]}: InternVL2 $>$ Cambrian $>$ LLaVA-1.6 $>$ LLaVA-1.5. While the dataset size for InternVL2 is unknown, it is likely $>$~InternVL1.2 with 45M. However, dataset quality is equally important: In ~\cref{fig:model-size} \emph{[Right]}, models that are trained with 1B data are less reliable than Cambrian on 0.01B data. Compared to the carefully curated, diverse datasets used by the most reliable models, these large datasets are mostly web-crawled, weakly-cleaned, and auto-augmented with model-generated captions.
Overall, we find that larger models per family and models from families trained on more high-quality data are more reliable,  making diverse, curated training data a potential path towards more reliable models.

%%%%%%%%%%%% RELATED WORKS
\vspace{-.2\baselineskip}
\section{Related Works}

Below, we detail how \framework{}'s solutions for VLM prompt sensitivity analysis -- prompt variations, reliability score and calibrated metrics -- are distinct from prior work.

\vspace{-.2\baselineskip}
\para{Prompt variations and prompt sensitivity}
Prompt sensitivity is severely under-explored for VLMs.
Few~\cite{qi2023limitation,chen2023benchmarking,khan2024consistency,schiappa2022robustness} probe VLMs or Video-LMs with noisy, corrupted images or randomly shuffled characters and words in text prompts, but unlike \framework{}'s prompt variations these are not reflective of realistic variations that could stem from human users.
For LLMs, prompt variations include reformulations and semantic changes.
Reformulations range from punctuation marks or random word changes~\cite{voronov2024mind,gu-etal-2023-robustness} over sentence rephrases~\cite{sun2024evaluating,loya-etal-2023-exploring,gan-mori-2023-sensitivity,sclar2024quantifying} or chain-of-thought re-prompting~\cite{mizrahi2023state} to reformulations for the best model performance~\cite{lu-etal-2022-fantastically,mishra2023promptaid,gonen-etal-2023-demystifying}.
Semantic prompt changes~\cite{mizrahi2023state,elazar2021measuring,jang-etal-2022-becel,sun2024evaluating} focus on negations like \enquote{What is red?}\ vs.\ \enquote{What is not red?}~\cite{anschutz-etal-2023-correct,zhang2023beyond,hosseini-etal-2021-understanding}.
Both prompt reformulation and semantic prompt changes are investigated in \cite{jang-etal-2022-becel}, but its benchmark and analysis are limited to LLMs.

\vspace{-.2\baselineskip}
\para{Scores}
VLM prompt sensitivity is commonly evaluated with several scores:
\textit{Accuracy}~\cite{kil2024compbench, liu2023mmbench} for correctness,
\textit{certainty}~\cite{angelopoulos2021gentle,kostumov2024uncertainty,ye2024benchmarking} for prediction confidence,
and \textit{consistency}~\cite{hudson2019gqa,jang-etal-2022-becel,khan2024consistency} for agreement between prompts.
There are scores beyond accuracy, built specifically for negations~\cite{sellam2020bleurt,anschutz-etal-2023-correct}, but BLEURT~\cite{sellam2020bleurt} score predictions requires an additional BERT model and hence cannot holistically evaluate sensitivity.
Certainty~\cite{kumar2023conformal,angelopoulos2021gentle,ren2023robots,ovadia2019can} is also commonly used, which conveys model confidence in its own predictions.
To measure if a model reliably answers similar questions,~\cite{kabra2023evaluating} aggregates log-likelihood of outputs across prompts.
Closest in spirit to our reliability is~\cite{kostumov2024uncertainty}, as it also combines certainty and accuracy, but in a less interpretable manner and with one rather than two explicit guarantees.

\vspace{-.2\baselineskip}
\para{Calibration}
Calibrated scores enable fair comparisons with differing baseline-performances, \eg in medicine~\cite{huang2020tutorial,licher2018external} and robustness~\cite{hendrycks2018benchmarking,taori2020measuring}, but have not been considered for VLM evaluations and the previous scores.
The calibration step introduced in \framework{} is the first to enable VLM sensitivity comparisons across datasets and prompts.

\vspace{-.2\baselineskip}
\para{VLMs}
VLMs combine LLMs with vision encoders to integrate vision and language data, enabling high performance on a variety of tasks, \eg visual question answering and captioning. Earlier models, \eg CLIP~\cite{radford2021CLIP} and ALIGN~\cite{jia2021align}, use contrastive learning to align language and vision to one embedding space.
Additional alignment attempts employ self-attention or cross-attention mechanisms.
Models such as LLaVA~\cite{liu2024visual,liu2023improvedllava} and PaLI~\cite{chen2022pali} use self-attention by mixing image and text tokens but cannot handle high-resolution inputs. Models like Flamingo, Qwen-VL, Phi-3-vision, CogVLM, BLIP-2, InternVL2, Cambrian~\cite{alayrac2022flamingo,bai2023qwen,abdin2024phi,wang2023cogvlm,li2023blip2,Chen2024internvl,tong2024cambrian1fullyopenvisioncentric} use cross-attention to efficiently manage visual tokens, independent of image resolution.
In \framework{}, we analyze model families with varying input resolutions and from self- and cross-attention groups.

%%%%%%%%% CONCLUSION
\section{Conclusion}

With \framework{}, we introduce a framework to analyze sensitivity to prompt variations in VLMs that is built on three pillars: A comprehensive set of vision and language prompt variations, an interpretable reliability score, and a principled calibration step to compare multiple sensitivity metrics across datasets and prompts.
Hence, \framework{} enables direct sensitivity comparisons between 22 VLMs from different families across seven datasets and 11 prompt variations for the first time.
Our analysis shows that language prompt sensitivity is mirrored in the vision domain and that VLMs struggle especially with prompt meaning as they are most vulnerable to semantic prompt alterations.
We also observe that model reliability and consistency are well correlated, indicating no inherent trade-off when selecting the most reliable versus the most consistent model.
Furthermore, models from specific families like InternVL2 are more agnostic to prompt variations; and within the same family, larger models like InternVL2 40B perform better.
Also, we find indications that high-quality training data may alleviate VLM prompt sensitivity.
We publicly release \framework{} and our prompt sensitivity  datasets so that the community may leverage our framework and insights to develop more reliable and consistent models.

%%%%%%%%% LIMITATIONS
\para{Limitations}
Our prompt sensitivity framework enables several key insights, but is aimed at white box VLMs for which we can calculate certainty. Although other sensitivity metrics can analyze black box models, their comprehensive application would be prohibitively expensive due to the sheer number of prompt changes. Also due to cost, we have one less semantic vision variation than other variations, because photo-realistic semantic changes require significant work to the point of a new dataset creation.
As prompt sensitivity datasets are limited, evaluating new datasets require efforts similar to our cleaning and expansion.
Lastly, we focus on MC-VQA datasets because discrete answers can be updated for semantic prompt variations, enabling certainty scores. Datasets for generative tasks require additional LLMs to validate the tested model's responses, injecting more noise and bias. In sum, \framework{} addresses sensitivity analyses in the most cost effective and impartial manner.

{\footnotesize\medskip\noindent\textbf{Acknowledgements.} JS acknowledges support through IMPRS-IS.}

\clearpage
\renewcommand{\thetable}{A\arabic{table}}
\renewcommand{\thefigure}{A\arabic{figure}}
\renewcommand{\thesection}{A\arabic{section}}
\maketitlesupplementary

\section{Evaluated Vision Language Models}

In~\cref{tab:supp_models} we give an overview of all 22 evaluated VLMs and their model properties, which form the basis for our experimental evaluation in~\cref{sec:experiments}.
The 22 models come from the 7 VLM families LLaVA1.5, LLaVA1.6, Qwen-VL, CogVLM, CogVLM2, InternVL2 and Cambrian~\cite{liu2023improvedllava,liu2024llavanext,bai2023qwen,wang2023cogvlm,hong2024cogvlm2,Chen2024internvl,tong2024cambrian1fullyopenvisioncentric}.

\begin{table}[t]
\centering
\scalebox{0.5}{
\begin{tabular}{@{}l@{\ }l@{}rrrr@{\ }c@{}}
\toprule
\multirow{2}{*}[-.0em]{\textbf{VLM Name}} & \multirow{2}{*}[-.0em]{\textbf{Huggingface ID}} & \multicolumn{1}{c}{\textbf{Model}} & \multicolumn{1}{c}{\textbf{LLM}} & \multicolumn{1}{c}{\textbf{Input}} & \multicolumn{1}{c}{\textbf{Training}} & \multicolumn{1}{c}{\textbf{Data}} \\
  & & \multicolumn{1}{c}{\textbf{Size}} & \multicolumn{1}{c}{\textbf{Size}} & \multicolumn{1}{c}{\textbf{Resolution}} & \multicolumn{1}{c}{\textbf{Data}} & \multicolumn{1}{c}{\textbf{Cur.}} \\
\midrule
\midrule
LLaVA-1.5 7B      & liuhaotian/llava-v1.5-7b & 7.4B & 7.0B & 336$\times$336 & 1.2M & high \\
LLaVA-1.5 13B     & liuhaotian/llava-v1.5-13b & 13.4B & 13.0B & 336$\times$336 & 1.2M & high \\
\midrule
LLaVA-1.6 7B vic  & liuhaotian/llava-v1.6-vicuna-7b & 7.1B & 6.7B & 336$\times$336 & 1.6M & high\\
LLaVA-1.6 7B mis  & liuhaotian/llava-v1.6-mistral-7b & 7.3B & 7.0B & 336$\times$336 & 1.6M & high\\
LLaVA-1.6 13B vic & liuhaotian/llava-v1.6-vicuna-13b & 13.3B & 13.0B & 336$\times$336 & 1.6M & high\\
LLaVA-1.6 34B     & liuhaotian/llava-v1.6-34b & 34.7B & 34.4B & 336$\times$336 & 1.6M & high \\
\midrule
Qwen-VL           & Qwen/Qwen-VL & 9.6B & 7.7B & 224$\times$224 & 1400.0M & low\\
Qwen-VL Chat      & Qwen/Qwen-VL-Chat & 9.6B & 7.7B & 224$\times$224 & 1400.0M & low \\
\midrule
CogVLM Chat       & THUDM/cogvlm-chat-hf & \emph{d.o.} & 7.0B & \emph{d.o.} & 2250.0M & low \\
CogAgent          & THUDM/cogagent-vqa-hf & \emph{d.o.} & 7.0B & \emph{d.o.} & 1690.0M & low \\
CogVLM GG         & THUDM/cogvlm-grounding-generalist-hf & \emph{d.o.} & 7.0B & \emph{d.o.} & 1500.0M & low \\
\midrule
CogVLM2 19B       & THUDM/cogvlm2-llama3-chat-19B & 19.0B & 8.0B & 1344$\times$1344 & \emph{n.a.} & \emph{n.a.} \\
\midrule
InternVL2 1B      & OpenGVLab/InternVL2-1B & 0.8B & 0.5B & 448$\times$448 & $>$50M & \emph{n.a.}\\
InternVL2 2B      & OpenGVLab/InternVL2-2B & 2.5B & 2.1B & 448$\times$448 & $>$50M & \emph{n.a.}\\
InternVL2 4B      & OpenGVLab/InternVL2-4B & 4.1B & 3.8B & 448$\times$448 & $>$50M & \emph{n.a.}\\
InternVL2 8B      & OpenGVLab/InternVL2-8B & 8.1B & 7.7B & 448$\times$448 & $>$50M & \emph{n.a.}\\
InternVL2 26B     & OpenGVLab/InternVL2-26B & 25.5B & 19.9B & 448$\times$448 & $>$50M & \emph{n.a.}\\
InternVL2 40B     & OpenGVLab/InternVL2-40B & 40.1B & 34.4B & 448$\times$448 & $>$50M & \emph{n.a.}\\
\midrule
Cambrian 3B       & nyu-visionx/cambrian-phi3-3b & \emph{d.o.} & 3.3B & $~$384$\times$384 & 9.5M & high \\
Cambrian 8B       & nyu-visionx/cambrian-8b & \emph{d.o.} & 8.0B & $~$384$\times$384 & 9.5M & high \\
Cambrian 13B      & nyu-visionx/cambrian-13b & \emph{d.o.} & 13.0B & $~$384$\times$384 & 9.5M & high \\
Cambrian 34B      & nyu-visionx/cambrian-34b & \emph{d.o.} & 34.4B & $~$384$\times$384 & 9.5M & high \\
\bottomrule
\end{tabular}
}
\vspace{-0.2cm}
\caption{Summary of analyzed VLMs in \framework{} and their properties. \enquote{Data Cur.} means Data Curation. The huggingface ID was used in our evaluation code to load the respective models. VLM properties are listed according to the corresponding papers or their github repositories. Note that some information marked with \enquote{\emph{n.a.}} is not publicly available, while \enquote{\emph{d.o.}} indicates that the information is difficult to obtain. The training data estimate for InternVL2 is based on its predecessor InternVL 1.2 plus with 39.3M pretraining and 12M finetuning data. The data curation estimate is \enquote{high} if the training data is a balanced collection of curated or established datasets for specific tasks, and \enquote{low} for web-crawled, auto-annotated data.}
\vspace{-0.15cm}
\label{tab:supp_models}
\end{table}

\section{\framework{} Prompt Variations and Datasets}

Below we give more details on the generation of our prompt variations and datasets.
First, we describe our prompt variation taxonomy in greater detail.
Next, we provide the LLaMA3 prompts~\cite{lm3}, which were used to generate the language variations across all datasets.
Then, we give details on creating the semantic vision exchange variation (VS-E), which requires manual data annotation and results in 6 new datasets.
Finally, we describe how we cleaned the prompts in the MMBench dataset~\cite{liu2023mmbench} to enable our language prompt alterations with LLaMA3.

\subsection{Prompt Variation Taxonomy}

We define any change to a prompt's text or vision component as prompt variation, focusing on human-plausible variations that reformulate (same expected answer) or semantically change (changed expected answer) the prompt.
\framework{} proposes expanding \emph{semantic changes} to images, the remaining vision and language \emph{reformulations} in \framework{} are a subset of realistic, existing text and image \emph{perturbations}:
\vspace*{-.35\baselineskip}
\begin{align*}
&
\small
\text{reformulation }^\text{vision}_\text{language}  \subset \text{perturbation }^\text{image}_\text{text}  \subset \text{prompt variation} .
\end{align*}

\vspace*{-.35\baselineskip}
\noindent
We choose them to cover broad perturbation classes, established in prior works.
For language reformulations, we adopted several text perturbation classes from~\cite{schiappa2022robustness}:  AddText [LR-V Verbose], DropText [LR-C Concise], and TextStyle [LR-I Instruction]; other classes in~\cite{schiappa2022robustness} that yield syntactically incorrect texts are omitted in \framework{} because they are unlikely to be made by real-world users.
For vision reformulations, we adopt major image perturbation classes from image corruption benchmarks~\cite{hendrycks2018benchmarking,Michaelis2019BenchmarkingRobustnessObject,Kar20223dCommonCorruptions,schmalfuss2025RobustSpring}, using the classes from~\cite{hendrycks2018benchmarking}: Color [VR-L Lighting], Detail change [VR-B Blur], and Spatial changes [VR-R Rotation].
We selected three representative reformulations for language/vision as we also consider computational feasibility for VLM evaluations: 11 total variations on 7 datasets already lead to extensive 84 dataset evaluations \emph{per VLM}.
Therefore, three representatives of diverse perturbation classes maximize prompt variability under evaluation budget constraints.

\subsection{Language Variations: LLaMA3 Prompts}

For our language prompt variations, we use LLaMA3-70B~\cite{lm3} to alter the language component of the original prompts.
Below, we provide the complete prompts that were used with LLaMA3 to alter an original text prompt, the \texttt{<Prompt to alter>}, into the variants Instruction (LR-I), Concise (LR-C), Verbose (LR-V), Not (LS-N), Antonyms (LS-A) and More-Less (LS-M).

\para{LR-I -- Instructions}
\begin{Verbatim}[fontsize=\tiny]
Rewrite the given phrases into instructions. Keep the instruction short, and as 
close to the original sentence as possible. If the phrase is already an
instruction, keep the original phrase. Only return a single new instruction,
do NOT give additional explanations.

Q: "Are the two sofas the same color in the picture?"
A: "Determine if the two sofas are the same color in the picture."

Q: "Based on the image, how can fun and engaging toothbrush holders help
children develop better dental health habits?"
A: "Based on the image, identify how fun and engaging toothbrush holders can
help children develop better dental health habits."

Q: "Which pants' fit is more regular?"
A: "Determine which pants' fit is more regular."

Q: "In the picture, which direction is the teddy bear facing?"
A: "Identify the direction in which the teddy bear is facing in the picture."

Q: "<Prompt to alter>"
A: 
\end{Verbatim}

\para{LR-C -- Concise}
\begin{Verbatim}[fontsize=\tiny]
Rephrase the following question using fewer words. Make sure the meaning of the
question stays the same. Return the same question if it is not possible to use
fewer words. For your context, the question is about an image that is not provided
here. Ensure correct grammar. Only return a single new question, do NOT give
additional explanations. Use fewer words than the original question.

Q: "Which road is more paved?"
A: "More paved road?"

Q: "Are the two sofas the same color in the picture?"
A: "Are the sofas the same color?"

Q: "Which cloud is whiter?"
A: "Whiter cloud?"

Q: "Which jacket is more asymmetrical??"
A: "More asymmetrical jacket?"

Q: "<Prompt to alter>"
A: 
\end{Verbatim}

\para{LR-V -- Verbose}
\begin{Verbatim}[fontsize=\tiny]
Rephrase the following question to make it more verbose. Make sure the meaning of
the question stays the same, do NOT invent additional details. For your context,
the question is about an image that is not provided here. Ensure correct grammar.
Only return a single new question, do NOT give additional explanations.

Q: "<Prompt to alter>"
A: 
\end{Verbatim}

\para{LS-N -- Not}
\begin{Verbatim}[fontsize=\tiny]
Change the following questions to ask for the opposite by negating them with `not'.
Keep the question as close to the original as possible. Ensure correct grammar.
Return a single, complete question. Do NOT abbreviate parts of the question.
Do NOT give additional explanations.

Q: "Which road is more paved?"
A: "Which road is not more paved?"

Q: "Which umbrella is yellower?"
A: "Which umbrella is not yellower?"

Q: "Which cat is staring more directly at the camera?"
A: "Which cat is notstaring more directly at the camera?"

Q: "Which fence is burnter?"
A: "Which fence is not more burnt?"

Q: "<Prompt to alter>"
A: 
\end{Verbatim}

\para{LS-A -- Antonyms}
\begin{Verbatim}[fontsize=\tiny]
Change the following questions to ask for the opposite by using antonyms or words
with opposite meanings. Keep the question as close to the original as possible.
Ensure correct grammar. Return a single, complete question. Do NOT abbreviate parts
of the question. Do NOT give additional explanations.

Q: "Which road is more paved?"
A: "Which road is more unpaved?"

Q: "Which cat is staring more directly at the camera?"
A: "Which cat is looking more away from the camera?"

Q: "Which bike is more folded?"
A: "Which bike is more unfolded?"

Q: "In what direction is Chile from Peru?"
A: "In what direction is Peru from Chile?"

Q: "Which Python code can generate the content of the image?"
A: "Which Python code refuses generating the content of the image?"

Q: "What musn't Joe and Alice trade to each get what they want?"
A: "What musn't Joe and Alice trade to each get what they want?"

Q: "Which apple is closer to the camera?"
A: "Which apple is farther from the camera?"

Q: "Which bottle is more dented?"
A: "Which bottle is more intact?"

Q: "Which aluminum is more molten?"
A: "Which aluminum is more solid?"

Q: "<Prompt to alter>"
A: 
\end{Verbatim}

\newpage
\para{LS-M -- More-Less}
\begin{Verbatim}[fontsize=\tiny]
Change the following questions to ask for the opposite by switching the word `more'
for `less' or `fewer'. Keep the question as close to the original as possible.
Ensure correct grammar. Return a single, complete question. Do NOT abbreviate parts
of the question. Do NOT give additional explanations.

Q: "Which road is more paved?"
A: "Which road is less paved?"

Q: "Which umbrella is yellower?"
A: "Which umbrella is less yellow?"

Q: "Which bed is closer to the camera?"
A: "Which bed is less close to the camera?"

Q: "Which animal has longer hair?"
A: "Which animal has less long hair?"

Q: "Which cat is staring more directly at the camera?"
A: "Which cat is staring less directly at the camera?"

Q: "Which fence is burnter?"
A: "Which fence is less burnt?"

Q: "<Prompt to alter>"
A: 
\end{Verbatim}

\subsection{Vision Variations}

Not all datasets supports every semantic vision variations that are proposed in~\cref{tab:prompt-var-overview}.
Below, we describe how we create the visual semantic exchange variation (VS-E), and give an overview of the resulting datasets after this manual annotation process.

\para{VS-E -- Image Exchange}
The visual semantic exchange variation is a semantic change, and therefore needs to modify the correct answer by changing the image component of the prompt. 
The idea is as follows:
For a prompt with two images (left and right) and a question of the type \enquote{Which image is more ...?}, we exchange the image that has more of this attribute, \ie that \emph{wins} the comparison, for one that has much less of this attribute, \ie that \emph{loses} the comparison.
For example, in~\cref{tab:prompt-var-overview}, we ask \enquote{Which animal has longer fur?}.
Here, the alpaca on the left wins, because it has longer fur than the otter on the right.
Now we exchange the alpaca on the left with an animal that has less long fur than the otter on the right, such that the otter on the right becomes the winner.

\begin{table}[t]
\centering
\scalebox{0.8}{
\begin{tabular}{lr@{\ \ }r@{\ \ }r@{\ \ }r@{\ \ }r@{\ \ }r}
\toprule
& \textbf{M-S}\hphantom{\%} & \textbf{M-A}\hphantom{\%} & \textbf{V-S}\hphantom{\%} & \textbf{V-A}\hphantom{\%} & \textbf{NYU}\hphantom{\%} & \textbf{Fas}\hphantom{\%} \\
\midrule
\multicolumn{7}{c}{\textbf{Number of Data Samples}} \\
\midrule
\textbf{Initial Samples} & 197\hphantom{\%} & 597\hphantom{\%} & 940\hphantom{\%} & 470\hphantom{\%} & 1858\hphantom{\%} & 2435\hphantom{\%}\\
\midrule
\textbf{VS-E Filtering} & 56\hphantom{\%} & 112\hphantom{\%} & 108\hphantom{\%} & 71\hphantom{\%} & 385\hphantom{\%} & 114\hphantom{\%} \\
\textbf{VS-E Percentage} & 28.4\% & 18.8\% & 11.5\% & 15.1\% & 20.7\% & 4.7\% \\
\midrule
\multicolumn{7}{c}{\textbf{LLaVA-1.6 34B Error Rates}} \\
\midrule
\textbf{Initial} & 89.7 & 84.9 & 79.3 & 77.7 & 67.2 & 72.1 \\
\textbf{Filtered} & 87.5 & 85.7 & 86.1 & 92.5 & 70.4 & 71.9 \\
\bottomrule
\end{tabular}
}
\vspace{-0.2cm}
\caption{Dataset statistics for the datasets MIT-States (M-S), MIT-Attributes (M-A), VAW-States (V-S), VAW-Attributes (V-A), NYU-Depth~V2 (NYU) and  Fashionpedia (Fas)~\cite{isola2015MITStatesAttributes,pham2021VAW,silberman2012NYUDepthv2,jia2020fashionpedia} as used in \framework{}. The \emph{Initial} comparative prompts are as curated in CompBench~\cite{kil2024compbench}, which uses 20\% of the \emph{Initial} data for evaluations. Our additional data annotation to enable Vision Semantic Changes by Exchanging images (VS-E) also retains about 20\% of this data, on which we evaluate the prompt sensitivity with \framework{}. LLaVA-1.6 34B error rates on initial and filtered prompts.}
\vspace{-0.15cm}
\label{tab:supp_combench_stats}
\end{table}

In practice, we identify valid replacement images in the following way \emph{per dataset} $D=\{p_n\}_{n=1}^N$ that is composed of prompts $p=(I^w, I^l, T)$, which are triplets of a winning image $I^w$, a losing image $I^l$ and a text $T$:
\begin{enumerate}
    \item Identify unique question texts within the dataset, and collect all instances of prompts with those unique question texts. \Eg, collect all instances of prompts that ask \enquote{Which cloud is more white?}.
    Only keep prompts belonging to a unique question with at least two instances.
    \item From the prompts for one unique question text, collect all \emph{losing} images $I^l$, \eg all \enquote{less white clouds}.
    \item Then, per losing image, mark all losing images this image would win against as possible exchanges. \Eg mark dark gray clouds as possible exchanges when considering a lighter gray cloud.
    \item Go through the prompts for the unique question text again. Per prompt, check if the losing image has an annotated image it would win against. If so, exchange the winning image with the annotated image. Discard prompts where the losing image does not have an annotated exchange.
\end{enumerate}
This process also works if the role of winning and losing images is swapped. 
In the example from~\cref{tab:prompt-var-overview}, we would then exchange the losing image (the otter) with one that wins against the alpaca, \eg a mammoth.
We identify unique question texts in the datasets MIT-States (M-S), MIT-Attributes (M-A), VAW-States (V-S), VAW-Attributes (V-A), NYU-Depth~V2 (NYU) and Fashionpedia (Fas)~\cite{isola2015MITStatesAttributes,pham2021VAW,silberman2012NYUDepthv2,jia2020fashionpedia}, and manually label potential exchange images per dataset.

\begin{figure*}
   \centering
   \includegraphics[width=.32\linewidth]{plots/reliab_0.5.pdf}
   \includegraphics[width=.32\linewidth]{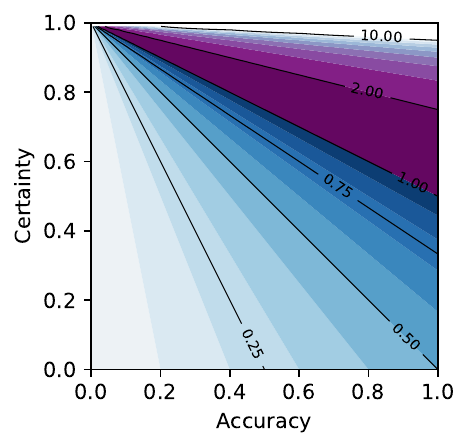}
   \includegraphics[width=.32\linewidth]{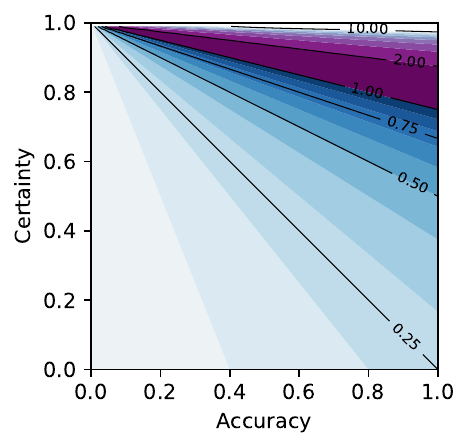}
   \caption{
   Mapping from certainty and accuracy to reliability score with $\text{acc}_\text{rand}=0.5$ \textit{[Left]} vs. uncertainty-aware accuracy UAcc with $|\mathcal{P}|=4$ \textit{[Middle]} and $|\mathcal{P}|=16$ \textit{[Right]}. Our reliability score provides two guarantees (one for accuracy and one for certainty) that can be directly seen from any score except 0. Uacc provides at most one guarantee for either accuracy or certainty. Further, its maximum changes with the number of answers per prompt $|\mathcal{P}|$, generating scores that are incomparable across datasets.
   }
   \label{fig:Rob-vs-Uacc}
\end{figure*}

\para{Dataset Statistics after VS-E}
The image exchange process described above reduces the number of available samples, because prompts are discarded if their text does not have multiple occurrences or if no suitable exchange image is found.
In~\cref{tab:supp_combench_stats} we show the number of initial prompts, and how many prompts are retained after discarding those whose images cannot be exchanged.
We retain about 20\% of the prompts, which is similar to the evaluation of those datasets in~\cite{kil2024compbench}, which randomly splits these datasets into 20\% test and 80\% training data.

We also report accuracies for LLaVA-1.6 34B on the original and filtered datasets.
They are generally well aligned, except for VAW-derived data.
For V-S and V-A, the filtered datasets are slightly easier, as filtering for exchangeable questions VS-E discards rare, difficult questions with single occurrences, which are more prominent on VAW.
We refrained from further filtering on these two datasets to maintain adequate sample size for subsequent evaluations.
Note that this does not affect insights from PARC, as all models are evaluated on the same data.

\subsection{MMBench Data Cleaning Protocol}

We use the prompts of the MMBench~\cite{liu2023mmbench} val split to create multiple prompt variations within \framework{}.
Because MMBench is composed from multiple data sources, the prompts do not follow a uniform structure.
To enable a consistent modification with our prompt variation framwork and ensure that the original prompts are sensical, we manually clean the 382 unique text prompts using the following steps:
\begin{enumerate}[label=\arabic*.,topsep=0.25\baselineskip]
\setlength{\itemsep}{.25\baselineskip}%
\setlength{\parskip}{0pt}%
\setlength{\topsep}{0pt}
\setlength{\labelindent}{0pt}%
    \item Remove unrelated sentence parts with convoluting information that is irrelevant to the question and its answer.
    \item Unify multiple sentences into one.
    \item Rephrase prompts to grammatically correct questions.
    \item Unify sentence capitalization.
    \item Where possible, remove any answer-leakages from questions.
    \item Remove questions that do not support negations via antonym replacement.
\end{enumerate}
For steps 2 and 3, we use LLaMA3-70B~\cite{lm3} to generate an initial set of proposed questions which are manually checked and corrected where necessary.
The following prompt is used for the initial question rephrasing:
\begin{Verbatim}[fontsize=\tiny]
Rewrite the given phrases into questions. Keep the question as close to the
original sentence as possible. If the phrase is already a question, keep the
original question. Ensure correct grammar. Only return a single new question,
do NOT give additional explanations.

Q: "where is the bike?"
A: "Where is the bike?"

Q: "What direction is India in Kyrgyzstan?"
A: "In what direction is India from Kyrgyzstan?"

Q: "Based on the image, how can fun and engaging toothbrush holders help
children develop better dental health habits?"
A: "Based on the image, how can fun and engaging toothbrush holders help
children develop better dental health habits?"

Q: "Complete the sentence.
The African elephant is the () land animal in the world."
A: "How do you complete the following sentence? `The African elephant is
the () land animal in the world.'"

Q: "Two magnets are places as shown. Will these magnets attract or repel
each other?"
A: "If the two magnets are placed as shown, will they attract or repel
each other?"

Q: "<Prompt to Rephrase>"
A: 
\end{Verbatim}
In step 5, an example of removing answer-leakages from the following original prompt:
\begin{Verbatim}[fontsize=\footnotesize]
Based on the image, what does the dog's
behavior of jumping and playing Frisbee
indicate about its well-being?
(A) The dog is participating in a
    professional Frisbee competition.
(B) The dog is engaged in physical activity,
    promoting its health and well-being.
(C) The dog is attempting to catch 
    a bird in mid-air.
(D) The dog is bored and looking for
    something to do. 
\end{Verbatim}
This text question invalidates answers \texttt{C} and \texttt{D} without even seeing the image, because it mentions the dog interacts with a frisbee. 
After removal, the question becomes \texttt{Based on the image, what does the dog do?}

\section{Analysis of \framework{}'s Reliability Score}
\label{supp.sec:reliability score}

\begin{figure*}
   \centering
   \begin{tabular}{@{}c@{}c@{}c@{}}
    & \hspace{2.4cm} Accuracy \textbf{Before Calibration} & \hspace{2.4cm} Accuracy \textbf{After Calibration}
    \\
    \multirow{1}{*}[1.8cm]{\rotatebox[origin=c]{90}{ MMBench}} 
    & \multicolumn{2}{c}{\includegraphics[width=.96\linewidth]{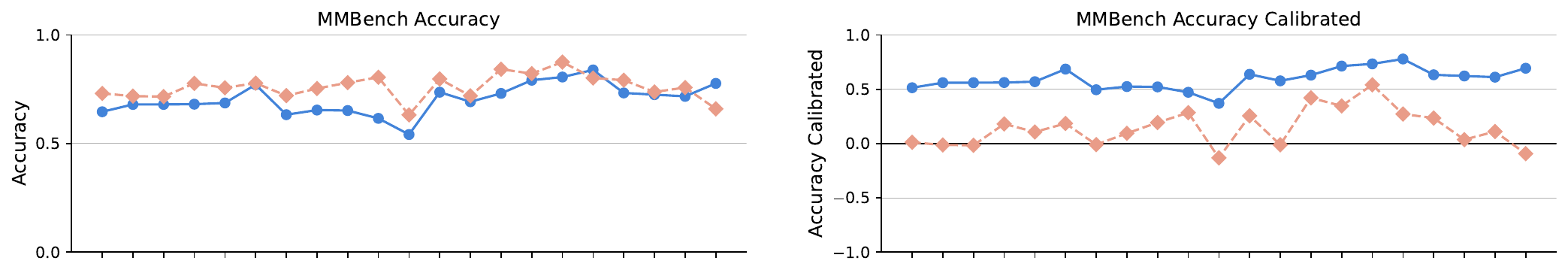}}
    \\
    \multirow{1}{*}[2.3cm]{\rotatebox[origin=c]{90}{NYU-Depth V2}}
    & \multicolumn{2}{c}{\includegraphics[width=.96\linewidth]{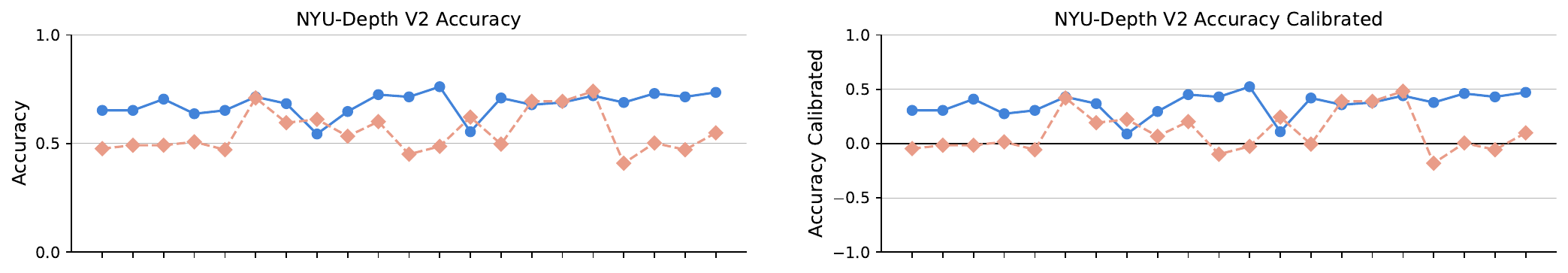}}
    \\
    \multirow{1}{*}[3.6cm]{\rotatebox[origin=c]{90}{Comparative Datasets}}
    & \multicolumn{2}{c}{\includegraphics[width=.96\linewidth]{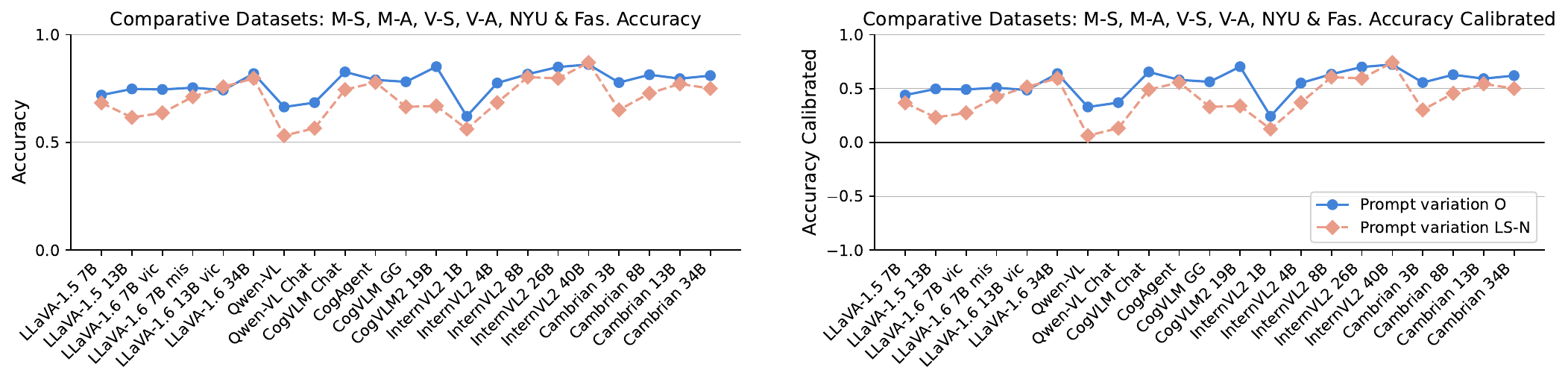}}
    \\
   \end{tabular}
    \vspace{-0.35cm}
   \caption{
   Effect of calibration, extension of~\cref{fig:calib_MMBench_CompBench} for all methods.
   Accuracy before \textit{[Left]} vs.\ after calibration \textit{[Right]} for original \textcolor{plotblue}{\textit{[Blue]}} and negated \textcolor{plotorange}{\textit{[orange]}} prompts. Only calibration aligns the ordering -- \textcolor{plotblue}{original} over \textcolor{plotorange}{negation} -- on the imbalanced MMBench~\cite{liu2023mmbench} dataset with the balanced NYU-Depth V2 dataset~\cite{silberman2012NYUDepthv2} and the average over all balanced comparative datasets~\cite{kil2024compbench}.
   In comparison to \cref{fig:calib_MMBench_CompBench}, this plot shows results for all methods and also the averages over all six balanced datasets in the last row.
   }
   \label{fig:supp_calib_MMBench_CompBench}
\end{figure*}

\begin{table*}[t]
\centering
\scalebox{0.42}{
\begin{tabular}{lll@{\quad\quad\quad}rrrrrrrr@{\quad\quad\quad}rrrrrrrr@{\quad\quad\quad}rrrrrrrr@{\quad\quad\quad}rrrrrrrr}
\toprule
& & & \multicolumn{8}{c}{\textbf{Reliability Calibrated}} & \multicolumn{8}{c}{\textbf{Accuracy Calibrated}} & \multicolumn{8}{c}{\textbf{Certainty Calibrated}} & \multicolumn{8}{c}{\textbf{Consistency Calibrated}} \\
\cmidrule(r{1cm}){4-11} \cmidrule(r{1cm}){12-19} \cmidrule(r{1cm}){20-27} \cmidrule(r){28-35}
\multicolumn{3}{l}{\textbf{Prompt Variation}} & M-S & M-A & V-S & V-A & NYU & Fas & MMB & \textbf{AVG} & M-S & M-A & V-S & V-A & NYU & Fas & MMB & \textbf{AVG} & M-S & M-A & V-S & V-A & NYU & Fas & MMB & \textbf{AVG} & M-S & M-A & V-S & V-A & NYU & Fas & MMB & \textbf{AVG} \\
\midrule
\midrule
\multicolumn{2}{l}{\textbf{Original}} &
\textbf{O}        & 0.52 & 0.12 & 0.24 & 0.23 & 0.13 & 0.32 &  0.48 & 0.29                    & 0.59 & 0.51 & 0.53 & 0.49 &  0.31 & 0.44 &  0.57 & 0.49                    & 0.83 & 0.26 & 0.49 & 0.46 & 0.41 & 0.64 & 0.73 & 0.54 &      \\
\midrule
\multirow{6}{*}[-.25em]{\rotatebox[origin=c]{90}{\textbf{Language}}}
& \multirow{3}{*}[0em]{\rotatebox[origin=c]{90}{\textbf{Ref.}}} &
\textbf{LR-I}     & 0.51 & 0.07 & 0.22 & 0.16 & 0.13 & 0.29 &  0.47 & 0.26                    & 0.60 & 0.41 & 0.46 & 0.44 &  0.30 & 0.43 &  0.57 & 0.46                    & 0.78 & 0.17 & 0.45 & 0.37 & 0.41 & 0.57 & 0.72 & 0.50 &     0.83 & 0.68 & 0.74 & 0.73 & 0.79 & 0.80 & 0.87 & 0.78 \\
&&
\textbf{LR-C}     & 0.41 & 0.13 & 0.17 & 0.15 & 0.11 & 0.28 &  0.45 & 0.24                    & 0.53 & 0.46 & 0.44 & 0.46 &  0.27 & 0.39 &  0.56 & 0.45                    & 0.69 & 0.27 & 0.42 & 0.35 & 0.40 & 0.56 & 0.70 & 0.49 &     0.78 & 0.73 & 0.71 & 0.79 & 0.64 & 0.63 & 0.84 & 0.73 \\
&&
\textbf{LR-V}     & 0.35 & 0.22 & 0.14 & 0.12 & 0.09 & 0.09 &  0.42 & 0.21                    & 0.54 & 0.57 & 0.34 & 0.32 &  0.27 & 0.34 &  0.53 & 0.42                    & 0.58 & 0.35 & 0.44 & 0.34 & 0.32 & 0.31 & 0.69 & 0.43 &     0.72 & 0.69 & 0.63 & 0.62 & 0.49 & 0.49 & 0.79 & 0.63 \\
\cmidrule(r){2-35}
&\multirow{3}{*}[0.em]{\rotatebox[origin=c]{90}{\textbf{Sem.}}} &
\textbf{LS-N}     & 0.39 & 0.06 & 0.22 & 0.18 & 0.04 & 0.25 &  0.05 & 0.17                    & 0.56 & 0.52 & 0.47 & 0.44 &  0.13 & 0.45 &  0.14 & 0.39                    & 0.61 & 0.11 & 0.41 & 0.35 & 0.23 & 0.48 & 0.38 & 0.37 &     0.58 & 0.50 & 0.42 & 0.44 & 0.06 & 0.37 & 0.06 & 0.35 \\
&&
\textbf{LS-A}     & 0.30 & 0.14 & 0.11 & 0.10 & 0.00 & 0.05 & -0.01 & 0.10                    & 0.48 & 0.42 & 0.39 & 0.25 & -0.03 & 0.24 & -0.08 & 0.24                    & 0.53 & 0.30 & 0.28 & 0.35 & 0.15 & 0.20 & 0.27 & 0.30 &     0.48 & 0.30 & 0.22 & 0.15 & -0.34 & 0.01 & -0.16 & 0.09 \\
&&
\textbf{LS-M}     & 0.15 & 0.09 & 0.21 & 0.14 & 0.01 & 0.17 &  \emph{n.a.} & 0.13                    & 0.39 & 0.57 & 0.47 & 0.38 & -0.02 & 0.38 &  \emph{n.a.} & 0.36                    & 0.35 & 0.15 & 0.44 & 0.31 & 0.16 & 0.40 & \emph{n.a.} & 0.30 &     0.40 & 0.42 & 0.30 & 0.28 & -0.26 & 0.28 &\emph{n.a.} & 0.24 \\
\midrule
\multirow{6}{*}[.5em]{\rotatebox[origin=c]{90}{\textbf{Vision}}}
&\multirow{3}{*}[0em]{\rotatebox[origin=c]{90}{\textbf{Ref.}}} &
\textbf{VR-B}     & 0.41 & 0.09 & 0.23 & 0.22 & 0.13 & 0.32 &  0.43 & 0.26                    & 0.49 & 0.43 & 0.49 & 0.50 &  0.31 & 0.45 &  0.54 & 0.46                    & 0.76 & 0.25 & 0.47 & 0.43 & 0.40 & 0.63 & 0.69 & 0.52 &     0.75 & 0.71 & 0.80 & 0.80 & 0.77 & 0.83 & 0.84 & 0.79 \\
&&
\textbf{VR-R}     & 0.48 & 0.21 & 0.23 & 0.29 & 0.08 & 0.13 &  0.38 & 0.26                    & 0.58 & 0.47 & 0.49 & 0.50 &  0.26 & 0.28 &  0.49 & 0.44                    & 0.79 & 0.42 & 0.48 & 0.57 & 0.35 & 0.38 & 0.65 & 0.52 &     0.79 & 0.70 & 0.68 & 0.72 & 0.53 & 0.56 & 0.74 & 0.67 \\
&&
\textbf{VR-L}     & 0.33 & 0.12 & 0.15 & 0.20 & 0.09 & 0.19 &  0.41 & 0.21                    & 0.43 & 0.42 & 0.34 & 0.46 &  0.26 & 0.33 &  0.52 & 0.39                    & 0.65 & 0.25 & 0.39 & 0.41 & 0.33 & 0.41 & 0.67 & 0.45 &     0.56 & 0.63 & 0.55 & 0.61 & 0.62 & 0.61 & 0.79 & 0.63 \\
\cmidrule(r){2-35}
&\multirow{3}{*}[0.75em]{\rotatebox[origin=c]{90}{\textbf{Sem.}}} &
\textbf{VS-S}     & 0.59 & 0.12 & 0.21 & 0.24 & 0.17 & 0.40 &  \emph{n.a.} & 0.29                    & 0.68 & 0.50 & 0.48 & 0.54 &  0.39 & 0.57 &  \emph{n.a.} & 0.53                    & 0.83 & 0.25 & 0.38 & 0.43 & 0.43 & 0.65 & \emph{n.a.} & 0.49 &     0.65 & 0.56 & 0.42 & 0.46 & 0.23 & 0.45 &\emph{n.a.} & 0.46 \\
&&
\textbf{VS-E}     & 0.13 & 0.19 & 0.04 & 0.11 & 0.16 & 0.16 &  \emph{n.a.} & 0.13                    & 0.28 & 0.41 & 0.30 & 0.31 &  0.38 & 0.30 &  \emph{n.a.} & 0.33                    & 0.41 & 0.45 & 0.15 & 0.34 & 0.42 & 0.39 & \emph{n.a.} & 0.36 &     0.13 & 0.09 & -0.04 & -0.04 & -0.14 & -0.02 &\emph{n.a.} & -0.00 \\
\bottomrule
\end{tabular}
}
\vspace{-0.2cm}
\caption{Most perturbing prompt variations in language and vision, averaged across models -- additional measurements for \cref{tab:promptvar-comp-reliab}. All scores are calibrated, with 1.0 being ideal model performance and 0.0 is random performance. The consistency is calculated between each varied and the original prompt. Most perturbing prompt per variation class is bold, and across variations underlined. High number indicates model robustness to a prompt variant. Prompt variation acronyms are from~\cref{tab:prompt-var-overview}. MMBench's non-comparative questions about single images neither support more-less reformulation (LS-M), nor image swaps or exchanges (VS-S, VS-E).}
\vspace{-0.15cm}
\label{tab:supp_promptvar-comp-reliab}
\end{table*}\begin{table*}[h!]
\centering
\scalebox{0.5}{
\begin{tabular}{l@{\quad}r@{\ \ }r@{\ \ }r@{\ \ }r@{\ \ }r@{\ \ }r@{\ \ }r@{\quad}r@{\quad\quad}r@{\ \ }r@{\ \ }r@{\ \ }r@{\ \ }r@{\ \ }r@{\ \ }r@{\quad}r@{\quad\quad}r@{\ \ }r@{\ \ }r@{\ \ }r@{\ \ }r@{\ \ }r@{\ \ }r@{\quad}r@{\quad\quad}r@{\ \ }r@{\ \ }r@{\ \ }r@{\ \ }r@{\ \ }r@{\ \ }r@{\quad}r@{}}
\toprule
& \multicolumn{8}{c}{\textbf{Reliability Calibrated}} & \multicolumn{8}{c}{\textbf{Accuracy Calibrated}} & \multicolumn{8}{c}{\textbf{Certainty Calibrated}} & \multicolumn{8}{c}{\textbf{Consistency Calibrated}} \\
\cmidrule(r{.5cm}){2-9} \cmidrule(r{.5cm}){10-17} \cmidrule(r{.5cm}){18-25} \cmidrule(r{.1cm}){25-32}
\textbf{Model} & M-S & M-A & V-S & V-A & NYU & Fas & MMB & \textbf{AVG} & M-S & M-A & V-S & V-A & NYU & Fas & MMB & \textbf{AVG} & M-S & M-A & V-S & V-A & NYU & Fas & MMB & \textbf{AVG} & M-S & M-A & V-S & V-A & NYU & Fas & MMB & \textbf{AVG} \\
\midrule
\midrule
LLaVA-1.5 7B      & 0.33 & 0.09 & 0.12 & 0.12 & 0.06 & 0.12 & 0.28 & 0.16              & 0.57 & 0.32 & 0.33 & 0.38 & 0.16 & 0.28 & 0.39 &  0.35 &                          0.56 & 0.29 & 0.34 & 0.29 & 0.29 & 0.35 & 0.56 & 0.38 &         0.62 & 0.40 & 0.48 & 0.52 & 0.29 & 0.42 & 0.60 & 0.48 \\
LLaVA-1.5 13B     & 0.27 & 0.17 & 0.17 & 0.14 & 0.08 & 0.11 & 0.31 & 0.18              & 0.46 & 0.47 & 0.37 & 0.42 & 0.18 & 0.35 & 0.40 &  0.38 &                          0.55 & 0.35 & 0.44 & 0.31 & 0.33 & 0.31 & 0.58 & 0.41 &         0.62 & 0.56 & 0.52 & 0.53 & 0.23 & 0.50 & 0.60 & 0.51 \\
\midrule
LLaVA-1.6 7B vic  & 0.34 & 0.09 & 0.15 & 0.16 & 0.12 & 0.11 & 0.30 & 0.18              & 0.52 & 0.35 & 0.34 & 0.46 & 0.26 & 0.24 & 0.40 &  0.37 &                          0.62 & 0.25 & 0.38 & 0.31 & 0.32 & 0.38 & 0.58 & 0.41 &         0.54 & 0.34 & 0.41 & 0.53 & 0.36 & 0.14 & 0.60 & 0.42 \\
LLaVA-1.6 7B mis  & 0.36 & 0.08 & 0.20 & 0.28 & 0.06 & 0.19 & 0.35 & 0.22              & 0.55 & 0.39 & 0.44 & 0.58 & 0.19 & 0.41 & 0.46 &  0.43 &                          0.65 & 0.20 & 0.46 & 0.46 & 0.28 & 0.42 & 0.63 & 0.44 &         0.56 & 0.56 & 0.48 & 0.62 & 0.31 & 0.59 & 0.62 & 0.54 \\
LLaVA-1.6 13B vic & 0.49 & 0.12 & 0.23 & 0.26 & 0.09 & 0.21 & 0.34 & 0.25              & 0.63 & 0.54 & 0.47 & 0.57 & 0.22 & 0.42 & 0.43 &  0.47 &                          0.76 & 0.22 & 0.44 & 0.45 & 0.33 & 0.39 & 0.63 & 0.46 &         0.76 & 0.57 & 0.57 & 0.65 & 0.30 & 0.54 & 0.62 & 0.57 \\
LLaVA-1.6 34B     & 0.53 & 0.25 & 0.32 & 0.36 & 0.15 & 0.25 & 0.44 & 0.33              & 0.67 & 0.67 & 0.55 & 0.65 & 0.35 & 0.39 & 0.49 &  0.54 &                          0.78 & 0.39 & 0.55 & 0.55 & 0.40 & 0.60 & 0.70 & 0.57 &         0.79 & 0.69 & 0.60 & 0.68 & 0.41 & 0.61 & 0.62 & 0.63 \\
\midrule
Qwen-VL           & 0.03 & 0.01 & 0.03 & 0.06 & 0.04 & 0.04 & 0.21 & 0.06              & 0.06 & 0.07 & 0.16 & 0.26 & 0.13 & 0.12 & 0.26 &  0.15 &                          0.33 & 0.18 & 0.25 & 0.25 & 0.24 & 0.28 & 0.51 & 0.29 &         0.00 & 0.14 & 0.11 & 0.24 & 0.16 & -0.06 & 0.42 & 0.14 \\
Qwen-VL Chat      & 0.11 & 0.05 & 0.09 & 0.15 & 0.04 & 0.05 & 0.25 & 0.11              & 0.29 & 0.35 & 0.30 & 0.34 & 0.14 & 0.16 & 0.35 &  0.27 &                          0.35 & 0.15 & 0.29 & 0.32 & 0.26 & 0.31 & 0.53 & 0.32 &         0.31 & 0.40 & 0.40 & 0.42 & 0.35 & 0.21 & 0.52 & 0.37 \\
\midrule
CogVLM Chat       & 0.48 & 0.12 & 0.32 & 0.22 & 0.09 & 0.34 & 0.26 & 0.26              & 0.67 & 0.58 & 0.61 & 0.54 & 0.20 & 0.47 & 0.36 &  0.49 &                          0.68 & 0.20 & 0.49 & 0.40 & 0.34 & 0.62 & 0.55 & 0.47 &         0.71 & 0.63 & 0.67 & 0.60 & 0.30 & 0.50 & 0.62 & 0.58 \\
CogAgent          & 0.42 & 0.12 & 0.21 & 0.14 & 0.10 & 0.32 & 0.23 & 0.22              & 0.58 & 0.54 & 0.56 & 0.47 & 0.26 & 0.47 & 0.34 &  0.46 &                          0.71 & 0.22 & 0.36 & 0.31 & 0.34 & 0.58 & 0.55 & 0.44 &         0.70 & 0.60 & 0.59 & 0.56 & 0.40 & 0.52 & 0.64 & 0.57 \\
CogVLM GG         & -0.10 & -0.07 & -0.26 & -0.23 & -0.04 & -0.15 & 0.02 & -0.12       & -0.24 & -0.22 & -0.16 & -0.22 & -0.01 & -0.26 & 0.09 & -0.15 &                    0.53 & 0.35 & 0.44 & 0.37 & 0.38 & 0.35 & 0.39 & 0.40 &         0.06 & -0.01 & -0.06 & -0.12 & -0.10 & -0.06 & 0.45 & 0.02 \\
\midrule
CogVLM2 19B       & 0.49 & 0.11 & 0.20 & 0.20 & 0.16 & 0.42 & 0.26 & 0.26              & 0.64 & 0.65 & 0.53 & 0.39 & 0.29 & 0.56 & 0.36 &  0.49 &                          0.71 & 0.18 & 0.38 & 0.48 & 0.45 & 0.69 & 0.52 & 0.49 &         0.68 & 0.68 & 0.51 & 0.38 & 0.30 & 0.58 & 0.39 & 0.50 \\
\midrule
InternVL2 1B      & 0.13 & 0.03 & 0.01 & 0.04 & 0.05 & 0.08 & 0.30 & 0.09              & 0.29 & 0.20 & 0.09 & 0.13 & 0.14 & 0.18 & 0.35 &  0.20 &                          0.36 & 0.16 & 0.18 & 0.28 & 0.25 & 0.21 & 0.53 & 0.28 &         0.32 & 0.40 & 0.31 & 0.18 & 0.15 & 0.26 & 0.54 & 0.31 \\
InternVL2 2B      & 0.23 & 0.15 & 0.11 & 0.11 & 0.09 & 0.14 & 0.37 & 0.17              & 0.39 & 0.49 & 0.44 & 0.33 & 0.25 & 0.31 & 0.47 &  0.38 &                          0.49 & 0.29 & 0.22 & 0.32 & 0.28 & 0.39 & 0.63 & 0.38 &         0.48 & 0.53 & 0.44 & 0.42 & 0.29 & 0.40 & 0.64 & 0.46 \\
InternVL2 4B      & 0.35 & 0.21 & 0.23 & 0.18 & 0.13 & 0.29 & 0.41 & 0.26              & 0.48 & 0.60 & 0.50 & 0.43 & 0.29 & 0.51 & 0.52 &  0.47 &                          0.69 & 0.36 & 0.44 & 0.39 & 0.36 & 0.53 & 0.67 & 0.49 &         0.67 & 0.61 & 0.50 & 0.59 & 0.34 & 0.52 & 0.64 & 0.55 \\
InternVL2 8B      & 0.51 & 0.23 & 0.30 & 0.24 & 0.11 & 0.34 & 0.48 & 0.32              & 0.65 & 0.68 & 0.62 & 0.46 & 0.34 & 0.55 & 0.59 &  0.56 &                          0.74 & 0.36 & 0.45 & 0.50 & 0.31 & 0.57 & 0.70 & 0.52 &         0.73 & 0.72 & 0.61 & 0.61 & 0.37 & 0.70 & 0.70 & 0.63 \\
InternVL2 26B     & 0.67 & 0.21 & 0.31 & 0.30 & 0.16 & 0.50 & 0.52 & 0.38              & 0.75 & 0.74 & 0.64 & 0.55 & 0.36 & 0.62 & 0.63 &  0.61 &                          0.86 & 0.28 & 0.46 & 0.55 & 0.42 & 0.74 & 0.73 & 0.58 &         0.83 & 0.77 & 0.69 & 0.70 & 0.47 & 0.69 & 0.72 & 0.70 \\
InternVL2 40B     & 0.69 & 0.20 & 0.33 & 0.33 & 0.17 & 0.48 & 0.57 & 0.40              & 0.77 & 0.73 & 0.78 & 0.59 & 0.38 & 0.65 & 0.65 &  0.65 &                          0.85 & 0.27 & 0.42 & 0.56 & 0.40 & 0.71 & 0.77 & 0.57 &         0.86 & 0.81 & 0.74 & 0.67 & 0.43 & 0.72 & 0.72 & 0.71 \\
\midrule
Cambrian 3B       & 0.48 & 0.16 & 0.18 & 0.11 & 0.10 & 0.23 & 0.39 & 0.24              & 0.61 & 0.53 & 0.42 & 0.37 & 0.22 & 0.51 & 0.49 &  0.45 &                          0.75 & 0.31 & 0.39 & 0.26 & 0.34 & 0.43 & 0.66 & 0.45 &         0.72 & 0.55 & 0.57 & 0.50 & 0.28 & 0.57 & 0.62 & 0.54 \\
Cambrian 8B       & 0.52 & 0.16 & 0.24 & 0.28 & 0.12 & 0.26 & 0.40 & 0.28              & 0.67 & 0.54 & 0.50 & 0.56 & 0.29 & 0.51 & 0.46 &  0.51 &                          0.74 & 0.29 & 0.45 & 0.46 & 0.35 & 0.47 & 0.66 & 0.49 &         0.80 & 0.69 & 0.54 & 0.57 & 0.43 & 0.57 & 0.60 & 0.60 \\
Cambrian 13B      & 0.51 & 0.20 & 0.23 & 0.20 & 0.11 & 0.25 & 0.38 & 0.27              & 0.64 & 0.61 & 0.53 & 0.48 & 0.25 & 0.47 & 0.46 &  0.49 &                          0.77 & 0.33 & 0.44 & 0.41 & 0.34 & 0.48 & 0.65 & 0.49 &         0.76 & 0.62 & 0.62 & 0.53 & 0.29 & 0.53 & 0.61 & 0.57 \\
Cambrian 34B      & 0.55 & 0.18 & 0.28 & 0.25 & 0.13 & 0.27 & 0.46 & 0.30              & 0.65 & 0.63 & 0.52 & 0.61 & 0.30 & 0.51 & 0.45 &  0.52 &                          0.84 & 0.28 & 0.53 & 0.42 & 0.37 & 0.51 & 0.71 & 0.52 &         0.84 & 0.77 & 0.59 & 0.71 & 0.41 & 0.59 & 0.60 & 0.64 \\
\bottomrule
\end{tabular}}
\vspace{-0.2cm}
\caption{Evaluation of model prompt sensitivity. Reliability, accuracy, certainty and consistency measured across models and datasets, and averaged over prompt variants. This expands~\cref{tab:model-comp-reliab} with the detailed numbers for accuracy and certainty. High numbers indicate high robustness. Note that CogVLM GG shows bad performance across all metrics, unless robustness, accuracy and consistency are calculated with the hightest-ranking logit score across answer choices in~\cref{tab:supp-model-comp-reliab_logits} rather than the direct VLM output.}
\vspace{-0.15cm}
\label{tab:supp-model-comp-reliab}
\end{table*}

\begin{table*}[h!]
\centering
\scalebox{0.5}{
\begin{tabular}{l@{\quad}r@{\ \ }r@{\ \ }r@{\ \ }r@{\ \ }r@{\ \ }r@{\ \ }r@{\quad}r@{\quad\quad}r@{\ \ }r@{\ \ }r@{\ \ }r@{\ \ }r@{\ \ }r@{\ \ }r@{\quad}r@{\quad\quad}r@{\ \ }r@{\ \ }r@{\ \ }r@{\ \ }r@{\ \ }r@{\ \ }r@{\quad}r@{\quad\quad}r@{\ \ }r@{\ \ }r@{\ \ }r@{\ \ }r@{\ \ }r@{\ \ }r@{\quad}r@{}}
\toprule
& \multicolumn{8}{c}{\textbf{Reliability Calibrated}} & \multicolumn{8}{c}{\textbf{Accuracy Calibrated}} & \multicolumn{8}{c}{\textbf{Certainty Calibrated}} & \multicolumn{8}{c}{\textbf{Consistency Calibrated}} \\
\cmidrule(r{.5cm}){2-9} \cmidrule(r{.5cm}){10-17} \cmidrule(r{.5cm}){18-25} \cmidrule(r{.1cm}){25-32}
\textbf{Model} & M-S & M-A & V-S & V-A & NYU & Fas & MMB & \textbf{AVG} & M-S & M-A & V-S & V-A & NYU & Fas & MMB & \textbf{AVG} & M-S & M-A & V-S & V-A & NYU & Fas & MMB & \textbf{AVG} & M-S & M-A & V-S & V-A & NYU & Fas & MMB & \textbf{AVG} \\
\midrule
\midrule
LLaVA-1.5 7B      & 0.33 & 0.09 & 0.12 & 0.12 & 0.06 & 0.12 & 0.28 & 0.16              & 0.57 & 0.32 & 0.33 & 0.38 & 0.16 & 0.28 & 0.39 &  0.35 &                          0.56 & 0.29 & 0.34 & 0.29 & 0.29 & 0.35 & 0.56 & 0.38 &         0.62 & 0.40 & 0.48 & 0.52 & 0.29 & 0.42 & 0.60 & 0.48 \\
LLaVA-1.5 13B     & 0.27 & 0.17 & 0.17 & 0.14 & 0.08 & 0.11 & 0.31 & 0.18              & 0.46 & 0.47 & 0.37 & 0.42 & 0.18 & 0.35 & 0.40 &  0.38 &                          0.55 & 0.35 & 0.44 & 0.31 & 0.33 & 0.31 & 0.58 & 0.41 &         0.62 & 0.56 & 0.52 & 0.53 & 0.23 & 0.50 & 0.60 & 0.51 \\
\midrule
LLaVA-1.6 7B vic  & 0.34 & 0.09 & 0.15 & 0.16 & 0.12 & 0.11 & 0.30 & 0.18              & 0.52 & 0.35 & 0.34 & 0.46 & 0.26 & 0.24 & 0.40 &  0.37 &                          0.62 & 0.25 & 0.38 & 0.31 & 0.32 & 0.38 & 0.58 & 0.41 &         0.54 & 0.34 & 0.41 & 0.53 & 0.36 & 0.14 & 0.60 & 0.42 \\
LLaVA-1.6 7B mis  & 0.36 & 0.08 & 0.20 & 0.28 & 0.06 & 0.19 & 0.35 & 0.22              & 0.55 & 0.39 & 0.44 & 0.58 & 0.19 & 0.41 & 0.46 &  0.43 &                          0.65 & 0.20 & 0.46 & 0.46 & 0.28 & 0.42 & 0.63 & 0.44 &         0.56 & 0.56 & 0.48 & 0.62 & 0.31 & 0.59 & 0.62 & 0.54 \\
LLaVA-1.6 13B vic & 0.49 & 0.12 & 0.23 & 0.26 & 0.09 & 0.21 & 0.34 & 0.25              & 0.63 & 0.54 & 0.47 & 0.57 & 0.22 & 0.42 & 0.43 &  0.47 &                          0.76 & 0.22 & 0.44 & 0.45 & 0.33 & 0.39 & 0.63 & 0.46 &         0.76 & 0.57 & 0.57 & 0.65 & 0.30 & 0.54 & 0.62 & 0.57 \\
LLaVA-1.6 34B     & 0.53 & 0.25 & 0.32 & 0.36 & 0.15 & 0.25 & 0.44 & 0.33              & 0.67 & 0.67 & 0.55 & 0.65 & 0.35 & 0.39 & 0.49 &  0.54 &                          0.78 & 0.39 & 0.55 & 0.55 & 0.40 & 0.60 & 0.70 & 0.57 &         0.79 & 0.69 & 0.60 & 0.68 & 0.41 & 0.61 & 0.62 & 0.63 \\
\midrule
Qwen-VL           & 0.03 & 0.01 & 0.03 & 0.06 & 0.04 & 0.04 & 0.21 & 0.06              & 0.06 & 0.07 & 0.16 & 0.26 & 0.13 & 0.12 & 0.26 &  0.15 &                          0.33 & 0.18 & 0.25 & 0.25 & 0.24 & 0.28 & 0.51 & 0.29 &         0.00 & 0.14 & 0.11 & 0.24 & 0.16 & -0.06 & 0.42 & 0.14 \\
Qwen-VL Chat      & 0.11 & 0.05 & 0.09 & 0.15 & 0.04 & 0.05 & 0.25 & 0.11              & 0.29 & 0.35 & 0.30 & 0.34 & 0.14 & 0.16 & 0.35 &  0.27 &                          0.35 & 0.15 & 0.29 & 0.32 & 0.26 & 0.31 & 0.53 & 0.32 &         0.31 & 0.40 & 0.40 & 0.42 & 0.35 & 0.21 & 0.52 & 0.37 \\
\midrule
CogVLM Chat       & 0.48 & 0.12 & 0.32 & 0.22 & 0.09 & 0.34 & 0.26 & 0.26              & 0.67 & 0.58 & 0.61 & 0.54 & 0.20 & 0.47 & 0.36 &  0.49 &                          0.68 & 0.20 & 0.49 & 0.40 & 0.34 & 0.62 & 0.55 & 0.47 &         0.71 & 0.63 & 0.67 & 0.60 & 0.30 & 0.50 & 0.62 & 0.58 \\
CogAgent          & 0.42 & 0.12 & 0.21 & 0.14 & 0.10 & 0.32 & 0.23 & 0.22              & 0.58 & 0.54 & 0.56 & 0.47 & 0.26 & 0.47 & 0.34 &  0.46 &                          0.71 & 0.22 & 0.36 & 0.31 & 0.34 & 0.58 & 0.55 & 0.44 &         0.70 & 0.60 & 0.59 & 0.56 & 0.40 & 0.52 & 0.64 & 0.57 \\
CogVLM GG         & -0.10 & -0.07 & -0.26 & -0.23 & -0.04 & -0.15 & 0.02 & -0.12       & -0.24 & -0.22 & -0.16 & -0.22 & -0.01 & -0.26 & 0.09 & -0.15 &                    0.53 & 0.35 & 0.44 & 0.37 & 0.38 & 0.35 & 0.39 & 0.40 &         0.06 & -0.01 & -0.06 & -0.12 & -0.10 & -0.06 & 0.45 & 0.02 \\
\midrule
CogVLM2 19B       & 0.49 & 0.11 & 0.20 & 0.20 & 0.16 & 0.42 & 0.26 & 0.26              & 0.64 & 0.65 & 0.53 & 0.39 & 0.29 & 0.56 & 0.36 &  0.49 &                          0.71 & 0.18 & 0.38 & 0.48 & 0.45 & 0.69 & 0.52 & 0.49 &         0.68 & 0.68 & 0.51 & 0.38 & 0.30 & 0.58 & 0.39 & 0.50 \\
\midrule
InternVL2 1B      & 0.13 & 0.03 & 0.01 & 0.04 & 0.05 & 0.08 & 0.30 & 0.09              & 0.29 & 0.20 & 0.09 & 0.13 & 0.14 & 0.18 & 0.35 &  0.20 &                          0.36 & 0.16 & 0.18 & 0.28 & 0.25 & 0.21 & 0.53 & 0.28 &         0.32 & 0.40 & 0.31 & 0.18 & 0.15 & 0.26 & 0.54 & 0.31 \\
InternVL2 2B      & 0.23 & 0.15 & 0.11 & 0.11 & 0.09 & 0.14 & 0.37 & 0.17              & 0.39 & 0.49 & 0.44 & 0.33 & 0.25 & 0.31 & 0.47 &  0.38 &                          0.49 & 0.29 & 0.22 & 0.32 & 0.28 & 0.39 & 0.63 & 0.38 &         0.48 & 0.53 & 0.44 & 0.42 & 0.29 & 0.40 & 0.64 & 0.46 \\
InternVL2 4B      & 0.35 & 0.21 & 0.23 & 0.18 & 0.13 & 0.29 & 0.41 & 0.26              & 0.48 & 0.60 & 0.50 & 0.43 & 0.29 & 0.51 & 0.52 &  0.47 &                          0.69 & 0.36 & 0.44 & 0.39 & 0.36 & 0.53 & 0.67 & 0.49 &         0.67 & 0.61 & 0.50 & 0.59 & 0.34 & 0.52 & 0.64 & 0.55 \\
InternVL2 8B      & 0.51 & 0.23 & 0.30 & 0.24 & 0.11 & 0.34 & 0.48 & 0.32              & 0.65 & 0.68 & 0.62 & 0.46 & 0.34 & 0.55 & 0.59 &  0.56 &                          0.74 & 0.36 & 0.45 & 0.50 & 0.31 & 0.57 & 0.70 & 0.52 &         0.73 & 0.72 & 0.61 & 0.61 & 0.37 & 0.70 & 0.70 & 0.63 \\
InternVL2 26B     & 0.67 & 0.21 & 0.31 & 0.30 & 0.16 & 0.50 & 0.52 & 0.38              & 0.75 & 0.74 & 0.64 & 0.55 & 0.36 & 0.62 & 0.63 &  0.61 &                          0.86 & 0.28 & 0.46 & 0.55 & 0.42 & 0.74 & 0.73 & 0.58 &         0.83 & 0.77 & 0.69 & 0.70 & 0.47 & 0.69 & 0.72 & 0.70 \\
InternVL2 40B     & 0.69 & 0.20 & 0.33 & 0.33 & 0.17 & 0.48 & 0.57 & 0.40              & 0.77 & 0.73 & 0.78 & 0.59 & 0.38 & 0.65 & 0.65 &  0.65 &                          0.85 & 0.27 & 0.42 & 0.56 & 0.40 & 0.71 & 0.77 & 0.57 &         0.86 & 0.81 & 0.74 & 0.67 & 0.43 & 0.72 & 0.72 & 0.71 \\
\midrule
Cambrian 3B       & 0.48 & 0.16 & 0.18 & 0.11 & 0.10 & 0.23 & 0.39 & 0.24              & 0.61 & 0.53 & 0.42 & 0.37 & 0.22 & 0.51 & 0.49 &  0.45 &                          0.75 & 0.31 & 0.39 & 0.26 & 0.34 & 0.43 & 0.66 & 0.45 &         0.72 & 0.55 & 0.57 & 0.50 & 0.28 & 0.57 & 0.62 & 0.54 \\
Cambrian 8B       & 0.52 & 0.16 & 0.24 & 0.28 & 0.12 & 0.26 & 0.40 & 0.28              & 0.67 & 0.54 & 0.50 & 0.56 & 0.29 & 0.51 & 0.46 &  0.51 &                          0.74 & 0.29 & 0.45 & 0.46 & 0.35 & 0.47 & 0.66 & 0.49 &         0.80 & 0.69 & 0.54 & 0.57 & 0.43 & 0.57 & 0.60 & 0.60 \\
Cambrian 13B      & 0.51 & 0.20 & 0.23 & 0.20 & 0.11 & 0.25 & 0.38 & 0.27              & 0.64 & 0.61 & 0.53 & 0.48 & 0.25 & 0.47 & 0.46 &  0.49 &                          0.77 & 0.33 & 0.44 & 0.41 & 0.34 & 0.48 & 0.65 & 0.49 &         0.76 & 0.62 & 0.62 & 0.53 & 0.29 & 0.53 & 0.61 & 0.57 \\
Cambrian 34B      & 0.55 & 0.18 & 0.28 & 0.25 & 0.13 & 0.27 & 0.46 & 0.30              & 0.65 & 0.63 & 0.52 & 0.61 & 0.30 & 0.51 & 0.45 &  0.52 &                          0.84 & 0.28 & 0.53 & 0.42 & 0.37 & 0.51 & 0.71 & 0.52 &         0.84 & 0.77 & 0.59 & 0.71 & 0.41 & 0.59 & 0.60 & 0.64 \\
\bottomrule
\end{tabular}}
\vspace{-0.2cm}
\caption{Evaluation of model prompt sensitivity. Reliability, accuracy, certainty and consistency measured across models and datasets, and averaged over prompt variants. This expands~\cref{tab:model-comp-reliab} with the detailed numbers for accuracy and certainty. High numbers indicate high robustness. Note that CogVLM GG shows bad performance across all metrics, unless robustness, accuracy and consistency are calculated with the hightest-ranking logit score across answer choices in~\cref{tab:supp-model-comp-reliab_logits} rather than the direct VLM output.}
\vspace{-0.15cm}
\label{tab:supp-model-comp-reliab}
\end{table*}

\begin{table}[]
\centering
\scalebox{0.9}{
\begin{tabular}{@{}l@{\ \ }r@{\ \ }r@{\ \ }r@{\ \ }rl@{\ \ }r@{\ \ }r@{\ \ }r@{\ \ }r@{}}
\toprule
  & Rel & Acc & Cert & Con &   & Rel & Acc & Cert & Con \\
\midrule
\textbf{LR}    & 0.26 & 0.46 & 0.50 & 0.78 & \textbf{LS}    & 0.17 & 0.39 & 0.37 & 0.35 \\
\textbf{VR}    & 0.21 & 0.39 & 0.45 & 0.63 & \textbf{VR}    & 0.21 & 0.39 & 0.45 & 0.63 \\
\textbf{LR+VR} & \underline{0.18} & \underline{0.37} & \underline{0.41} & \underline{0.59} & \textbf{LS+VR} & \underline{0.11} & \underline{0.31} & \underline{0.32} & \underline{0.26} \\
\midrule
\textbf{LR}    & 0.26 & \underline{0.46} & 0.50 & 0.78 & \textbf{LS}    & 0.17 & 0.39 & \underline{0.37} & \underline{0.35} \\
\textbf{VS}    & 0.29 & 0.53 & 0.49 & 0.46 & \textbf{VS}    & 0.29 & 0.53 & 0.49 & 0.46 \\
\textbf{LR+VS} & \underline{0.24} & \underline{0.46} & \underline{0.45} & \underline{0.42} & \textbf{LS+VS} & \underline{0.16} & \underline{0.35} & 0.38 & 0.42 \\
\bottomrule
\end{tabular}
}
    \caption{Effect of combining vision and language prompt variations on reliability, accuracy, certainty, and consistency, when averaged across all datasets. Expands \cref{tab:promptvar-comp-reliab} and \cref{tab:supp_promptvar-comp-reliab}.}
    \label{tab:supp_promptvar_combinations}

    \vspace{-\baselineskip}
\end{table}

\para{Calibration}
For calibration, the reliability should be 0 for $\mathit{acc}\!=\!\mathit{acc}_\mathit{rand}$, independent of $\mathit{cert}$.
This condition implicitly calibrates the vanilla definition $\mathit{rel}\!=\!(2\!\cdot\!\mathit{acc}\!-\!1)\!\cdot\!\mathit{cert}$ to $\mathit{acc}_\mathit{rand}\!=\!0.5$.
Thus to calibrate, we generalize the reliability definition to $\mathit{rel}\!=\!(2\!\cdot\!\mathit{acc}^m\!-\!1)\!\cdot\!\mathit{cert}$ and solve $0\!=\!(2\cdot \mathit{acc}_\mathit{rand}^m - 1)$, obtaining $m\!=\!\frac{\log(1/2)}{\log(\mathit{acc}_\mathit{rand})}\!=\!\frac{2}{\log(1/\mathit{acc}_\mathit{rand})}$.
For $\mathit{acc}_\mathit{rand}\!=\!0.5$ this translates to $m\!=\!1$ and yields the vanilla reliability, validating our generalization.

\para{Comparison to Other Scores}
The reliability score we propose with \framework{} is not the first to combine accuracy and uncertainty into a unified number.
Hence, we compare the interpretability of our reliability against the uncertainty-aware accuracy UAacc from \cite{kostumov2024uncertainty}.
UAcc is introduced within a framework that uses conformal prediction~\cite{angelopoulos2021gentle,ye2024benchmarking}.
Therefore, its certainty notion also builds on the set sizes of the set of all possible answers $\mathcal{P}$ and the prediction set $\mathcal{C}$:
\begin{align}
    \label{equ:Uacc}
    \text{UAcc} &= \frac{\text{acc}}{|\mathcal{C}|} \sqrt{|\mathcal{P}|} \\
    &= \frac{\text{acc}}{(1-\text{cert})\cdot(|\mathcal{P}|-1)+1} \sqrt{|\mathcal{P}|}.
\end{align}
UAcc takes values in $[0,\sqrt{|\mathcal{P}|}]$, where the maximum exceeds 1 and depends on the number of possible answers $|\mathcal{P}|$.
This makes it impossible to compare UAcc scores across datasets with varying answer numbers, because their score maximum is already inconsistent.
\cref{fig:Rob-vs-Uacc} compares the mapping from certainty and accuracy to UAcc with $|\mathcal{P}|=4$ and $|\mathcal{P}|=16$ to our reliability score.
And even within the same number of answers $|\mathcal{P}|$, UAcc does not offer truly meaningful built-in guarantees about the individual accuracies and certainties that can be seen from a single score at a glance: Any given score maps to the full spectrum of possible accuracies or certainties, \eg for $|\mathcal{P}|=4$, $\text{Uacc}\leq0.5$ may represent any certainty and $\text{Uacc}\geq 0.5$ may represent any accuracy.
For our reliability, only $\text{rel}=0$ is similarly indescriptive, while all other values come with the following two guarantees for certainty \emph{and} accuracy.
\begin{equation}
    \label{equ:supp_guarant}
    \text{cert} \geq |\text{rel}|, \quad \quad
    \text{acc}_\text{calib}
    \begin{cases}
    \geq \text{rel} & \quad \text{for acc}_\text{calib} > 0 \\
    \leq \text{rel} & \quad \text{for acc}_\text{calib} < 0
    \end{cases}.
\end{equation}

\section{\framework{}'s Calibration: Demonstration}

Our explanation of \framework{}'s calibration step in~\cref{fig:calib_MMBench_CompBench} was limited to a few selected methods. 
Therefore, we show the same evaluation but for all 22 evaluated methods in~\cref{fig:supp_calib_MMBench_CompBench}.
In addition to the calibration results on single datasets, we further show that the expected trend -- VLMs perform better on original than negated prompts -- holds across all 6 balanced datasets, and is not limited to NYU-Depth V2~\cite{silberman2012NYUDepthv2}.

\begin{table*}[]
\centering
\scalebox{0.52}{
\begin{tabular}{l@{\quad}r@{\ \ }r@{\ \ }r@{\ \ }r@{\ \ }r@{\ \ }r@{\ \ }r@{\quad}r@{\quad\quad}r@{\ \ }r@{\ \ }r@{\ \ }r@{\ \ }r@{\ \ }r@{\ \ }r@{\quad}r@{\quad\quad}r@{\ \ }r@{\ \ }r@{\ \ }r@{\ \ }r@{\ \ }r@{\ \ }r@{\quad}r@{\quad\quad}r@{\ \ }r@{\ \ }r@{\ \ }r@{\ \ }r@{\ \ }r@{\ \ }r@{\quad}r@{}}
\toprule
& \multicolumn{8}{c}{\textbf{Reliability Calibrated}} & \multicolumn{8}{c}{\textbf{Accuracy Calibrated}} & \multicolumn{8}{c}{\textbf{Certainty Calibrated}} & \multicolumn{8}{c}{\textbf{Consistency Calibrated}} \\
\cmidrule(r{.5cm}){2-9} \cmidrule(r{.5cm}){10-17} \cmidrule(r{.5cm}){18-25} \cmidrule(r{.1cm}){25-32}
\textbf{Model} & M-S & M-A & V-S & V-A & NYU & Fas & MMB & \textbf{AVG} & M-S & M-A & V-S & V-A & NYU & Fas & MMB & \textbf{AVG} & M-S & M-A & V-S & V-A & NYU & Fas & MMB & \textbf{AVG} & M-S & M-A & V-S & V-A & NYU & Fas & MMB & \textbf{AVG} \\
\midrule
\midrule
LLaVA-1.5 7B      & 0.33 & 0.09 & 0.12 & 0.12 & 0.06 & 0.12 & 0.28 & 0.16           & 0.57 & 0.32 & 0.33 & 0.38 & 0.16 & 0.28 & 0.39 & 0.35 &                  0.56 & 0.29 & 0.34 & 0.29 & 0.29 & 0.35 & 0.56 & 0.38 &          0.62 & 0.40 & 0.48 & 0.52 & 0.29 & 0.42 & 0.60 & 0.48 \\
LLaVA-1.5 13B     & 0.27 & 0.17 & 0.17 & 0.14 & 0.08 & 0.11 & 0.31 & 0.18           & 0.46 & 0.47 & 0.37 & 0.42 & 0.18 & 0.35 & 0.42 & 0.38 &                  0.55 & 0.35 & 0.44 & 0.31 & 0.33 & 0.31 & 0.58 & 0.41 &          0.62 & 0.56 & 0.52 & 0.53 & 0.23 & 0.50 & 0.60 & 0.51 \\
\midrule
LLaVA-1.6 7B vic  & 0.34 & 0.09 & 0.15 & 0.16 & 0.12 & 0.11 & 0.30 & 0.18           & 0.52 & 0.35 & 0.34 & 0.46 & 0.26 & 0.24 & 0.41 & 0.37 &                  0.62 & 0.25 & 0.38 & 0.31 & 0.32 & 0.38 & 0.58 & 0.41 &          0.54 & 0.34 & 0.41 & 0.53 & 0.36 & 0.14 & 0.59 & 0.42 \\
LLaVA-1.6 7B mis  & 0.36 & 0.08 & 0.20 & 0.28 & 0.06 & 0.19 & 0.35 & 0.22           & 0.55 & 0.39 & 0.44 & 0.58 & 0.19 & 0.41 & 0.46 & 0.43 &                  0.65 & 0.20 & 0.46 & 0.46 & 0.28 & 0.42 & 0.63 & 0.44 &          0.56 & 0.56 & 0.48 & 0.62 & 0.31 & 0.59 & 0.62 & 0.54 \\
LLaVA-1.6 13B vic & 0.49 & 0.12 & 0.23 & 0.26 & 0.09 & 0.21 & 0.34 & 0.25           & 0.63 & 0.54 & 0.47 & 0.57 & 0.22 & 0.42 & 0.44 & 0.47 &                  0.76 & 0.22 & 0.44 & 0.45 & 0.33 & 0.39 & 0.63 & 0.46 &          0.76 & 0.57 & 0.57 & 0.65 & 0.30 & 0.54 & 0.62 & 0.57 \\
LLaVA-1.6 34B     & 0.53 & 0.25 & 0.33 & 0.36 & 0.15 & 0.27 & 0.46 & 0.33           & 0.66 & 0.65 & 0.56 & 0.65 & 0.37 & 0.43 & 0.54 & 0.55 &                  0.78 & 0.39 & 0.55 & 0.55 & 0.40 & 0.60 & 0.70 & 0.57 &          0.81 & 0.70 & 0.67 & 0.67 & 0.50 & 0.66 & 0.65 & 0.67 \\
\midrule
Qwen-VL           & 0.15 & 0.04 & 0.07 & 0.07 & 0.06 & 0.07 & 0.23 & 0.10           & 0.38 & 0.18 & 0.28 & 0.27 & 0.22 & 0.21 & 0.34 & 0.27 &                  0.33 & 0.18 & 0.25 & 0.25 & 0.24 & 0.28 & 0.51 & 0.29 &          0.23 & 0.31 & 0.29 & 0.31 & 0.36 & 0.09 & 0.53 & 0.30 \\
Qwen-VL Chat      & 0.09 & 0.07 & 0.03 & 0.12 & 0.02 & 0.07 & 0.25 & 0.09           & 0.24 & 0.40 & 0.16 & 0.26 & 0.09 & 0.19 & 0.38 & 0.24 &                  0.35 & 0.15 & 0.29 & 0.32 & 0.26 & 0.31 & 0.53 & 0.32 &          0.24 & 0.39 & 0.11 & 0.23 & 0.19 & 0.19 & 0.51 & 0.27 \\
\midrule
CogVLM Chat       & 0.48 & 0.12 & 0.32 & 0.22 & 0.09 & 0.34 & 0.28 & 0.26           & 0.67 & 0.58 & 0.62 & 0.54 & 0.20 & 0.47 & 0.41 & 0.50 &                  0.68 & 0.20 & 0.49 & 0.40 & 0.34 & 0.62 & 0.55 & 0.47 &          0.71 & 0.63 & 0.67 & 0.61 & 0.30 & 0.50 & 0.62 & 0.58 \\
CogAgent          & 0.42 & 0.12 & 0.21 & 0.14 & 0.10 & 0.32 & 0.26 & 0.22           & 0.58 & 0.54 & 0.56 & 0.47 & 0.26 & 0.47 & 0.40 & 0.47 &                  0.71 & 0.22 & 0.36 & 0.31 & 0.34 & 0.58 & 0.55 & 0.44 &          0.70 & 0.60 & 0.59 & 0.56 & 0.40 & 0.52 & 0.64 & 0.57 \\
CogVLM GG         & 0.29 & 0.14 & 0.21 & 0.19 & 0.11 & 0.13 & 0.11 & 0.17           & 0.47 & 0.38 & 0.35 & 0.42 & 0.23 & 0.34 & 0.25 & 0.35 &                  0.53 & 0.35 & 0.44 & 0.37 & 0.38 & 0.35 & 0.39 & 0.40 &          0.55 & 0.45 & 0.33 & 0.43 & 0.26 & 0.35 & 0.50 & 0.41 \\
\midrule
CogVLM2 19B       & 0.54 & 0.11 & 0.24 & 0.24 & 0.18 & 0.47 & 0.26 & 0.29           & 0.72 & 0.67 & 0.64 & 0.45 & 0.33 & 0.62 & 0.37 & 0.54 &                  0.71 & 0.18 & 0.38 & 0.48 & 0.45 & 0.69 & 0.52 & 0.49 &          0.79 & 0.71 & 0.57 & 0.53 & 0.37 & 0.69 & 0.47 & 0.59 \\
\midrule
InternVL2 1B      & 0.13 & 0.03 & 0.01 & 0.04 & 0.05 & 0.08 & 0.30 & 0.09           & 0.29 & 0.20 & 0.09 & 0.13 & 0.14 & 0.18 & 0.40 & 0.20 &                  0.36 & 0.16 & 0.18 & 0.28 & 0.25 & 0.21 & 0.53 & 0.28 &          0.32 & 0.40 & 0.31 & 0.18 & 0.15 & 0.26 & 0.54 & 0.31 \\
InternVL2 2B      & 0.23 & 0.15 & 0.11 & 0.11 & 0.09 & 0.14 & 0.37 & 0.17           & 0.39 & 0.49 & 0.44 & 0.33 & 0.25 & 0.31 & 0.47 & 0.38 &                  0.49 & 0.29 & 0.22 & 0.32 & 0.28 & 0.39 & 0.63 & 0.38 &          0.48 & 0.53 & 0.44 & 0.42 & 0.29 & 0.40 & 0.64 & 0.46 \\
InternVL2 4B      & 0.35 & 0.21 & 0.21 & 0.19 & 0.12 & 0.27 & 0.41 & 0.25           & 0.47 & 0.59 & 0.46 & 0.40 & 0.27 & 0.47 & 0.52 & 0.46 &                  0.69 & 0.36 & 0.44 & 0.39 & 0.36 & 0.53 & 0.67 & 0.49 &          0.64 & 0.56 & 0.52 & 0.54 & 0.34 & 0.46 & 0.65 & 0.53 \\
InternVL2 8B      & 0.51 & 0.23 & 0.30 & 0.24 & 0.11 & 0.34 & 0.48 & 0.32           & 0.65 & 0.68 & 0.62 & 0.46 & 0.34 & 0.55 & 0.59 & 0.56 &                  0.74 & 0.36 & 0.45 & 0.50 & 0.31 & 0.57 & 0.70 & 0.52 &          0.73 & 0.72 & 0.61 & 0.61 & 0.37 & 0.70 & 0.69 & 0.63 \\
InternVL2 26B     & 0.67 & 0.21 & 0.31 & 0.30 & 0.16 & 0.50 & 0.52 & 0.38           & 0.75 & 0.74 & 0.64 & 0.55 & 0.36 & 0.62 & 0.62 & 0.61 &                  0.86 & 0.28 & 0.46 & 0.55 & 0.42 & 0.74 & 0.73 & 0.58 &          0.83 & 0.77 & 0.69 & 0.70 & 0.47 & 0.69 & 0.72 & 0.70 \\
InternVL2 40B     & 0.69 & 0.20 & 0.33 & 0.33 & 0.17 & 0.48 & 0.56 & 0.40           & 0.77 & 0.73 & 0.78 & 0.59 & 0.38 & 0.65 & 0.61 & 0.64 &                  0.85 & 0.27 & 0.42 & 0.56 & 0.40 & 0.71 & 0.77 & 0.57 &          0.86 & 0.81 & 0.74 & 0.67 & 0.43 & 0.72 & 0.68 & 0.70 \\
\midrule
Cambrian 3B       & 0.46 & 0.16 & 0.17 & 0.11 & 0.10 & 0.23 & 0.39 & 0.23           & 0.59 & 0.51 & 0.40 & 0.38 & 0.21 & 0.50 & 0.49 & 0.44 &                  0.75 & 0.31 & 0.39 & 0.26 & 0.34 & 0.43 & 0.66 & 0.45 &          0.70 & 0.51 & 0.54 & 0.51 & 0.26 & 0.54 & 0.64 & 0.53 \\
Cambrian 8B       & 0.52 & 0.15 & 0.23 & 0.28 & 0.12 & 0.26 & 0.40 & 0.28           & 0.67 & 0.52 & 0.48 & 0.56 & 0.30 & 0.50 & 0.47 & 0.50 &                  0.74 & 0.29 & 0.45 & 0.46 & 0.35 & 0.47 & 0.66 & 0.49 &          0.80 & 0.62 & 0.54 & 0.58 & 0.40 & 0.56 & 0.60 & 0.59 \\
Cambrian 13B      & 0.50 & 0.20 & 0.24 & 0.20 & 0.12 & 0.24 & 0.38 & 0.27           & 0.62 & 0.60 & 0.54 & 0.48 & 0.25 & 0.46 & 0.46 & 0.49 &                  0.77 & 0.33 & 0.44 & 0.41 & 0.34 & 0.48 & 0.65 & 0.49 &          0.78 & 0.66 & 0.62 & 0.48 & 0.29 & 0.52 & 0.61 & 0.57 \\
Cambrian 34B      & 0.55 & 0.18 & 0.28 & 0.26 & 0.13 & 0.27 & 0.46 & 0.30           & 0.64 & 0.64 & 0.51 & 0.62 & 0.31 & 0.50 & 0.49 & 0.53 &                  0.84 & 0.28 & 0.53 & 0.42 & 0.37 & 0.51 & 0.71 & 0.52 &          0.84 & 0.76 & 0.60 & 0.69 & 0.43 & 0.59 & 0.60 & 0.64 \\
\bottomrule
\end{tabular}}
\vspace{-0.2cm}
\caption{Evaluation of model prompt sensitivity with hightest logit score across answer options instead of raw VLM output. This mirrors~\cref{tab:supp-model-comp-reliab}. Note that the certainty values are unaffected by this change because certainty scores via conformal predictions already use logit scores. The trends and model behaviors remain almost unchanged, except for CogVLM GG, which now follows the trend of other CogVLM variants instead of underperforming significantly.}
\label{tab:supp-model-comp-reliab_logits}
\end{table*}

\section{Accompanying Results}

Due to space limitation, we reported only a subset of \framework{}'s metrics and potential evaluations in the main paper.
Below, we show extended metrics for the most perturbing prompts as well as additional analysis regarding the most prompt-agnostic VLMs.

\subsection{To Which Prompts are VLMs Sensitive?}

In~\cref{tab:supp_promptvar-comp-reliab} we show the additional detailed accuracy and certainty measures for our prompt type analysis in~\cref{tab:model-comp-reliab}.
The trends follow our reliability score.

\para{Combined Vision-Language Variations}
To test how combinations of prompt variations affect models, we select one variation 
per type (LR-I, LS-N, VR-L, VS-S) and evaluate the variation combinations LR-I/VR-L, LR-I/VS-S, LS-N/VR-L and LS-N/VS-S, resulting in 28 additional datasets per VLM.
In~\cref{tab:supp_promptvar_combinations}, we find that variation combinations behave like the worst variation or worse.
As combined variations follow the individual results, PARC's per-variation evaluation is a cost-effective strategy to avoid combination explosions.

\subsection{Which VLM is Most Prompt Agnostic?}

\begin{figure*}
   \centering
   \begin{tabular}{@{}c@{\ \ }c@{}c@{}c@{}c@{}}
        & \footnotesize\hspace{.5cm}\textbf{Language Only} & \footnotesize\hspace{.5cm}\textbf{Vision Only} & \footnotesize\hspace{.5cm}\textbf{Reformulation Only} & \footnotesize\hspace{.5cm}\textbf{Semantic Changes Only} \\
        \multirow{1}{*}[6.7em]{\rotatebox[origin=c]{90}{\footnotesize\textbf{Overview}}}
        & \includegraphics[width=.24\linewidth]{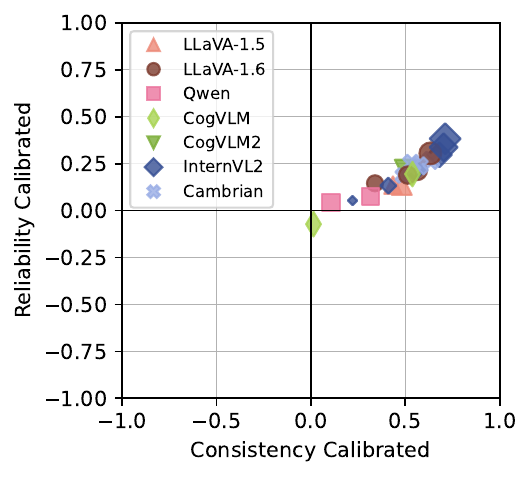}
         & \includegraphics[width=.24\linewidth]{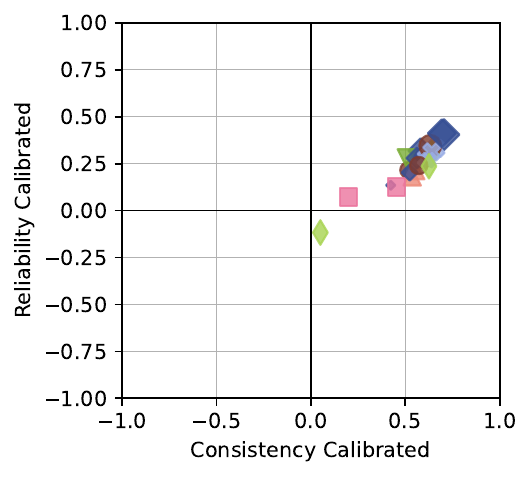}
         & \includegraphics[width=.24\linewidth]{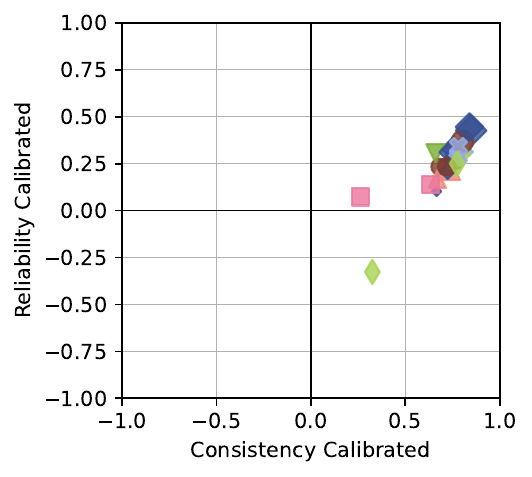}
         & \includegraphics[width=.24\linewidth]{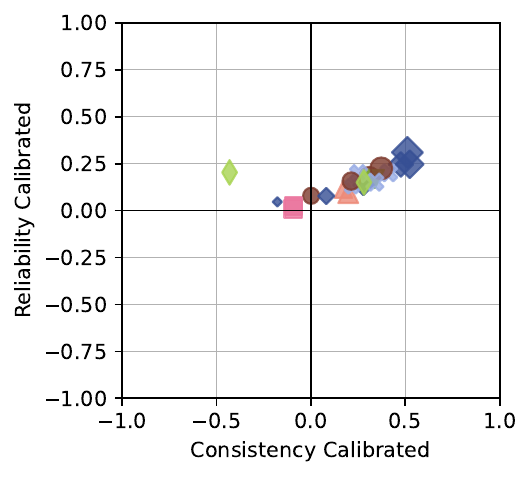}
         \\
        \multirow{1}{*}[7.5em]{\rotatebox[origin=c]{90}{\footnotesize\textbf{Detail View}}}
        &\includegraphics[width=.24\linewidth]{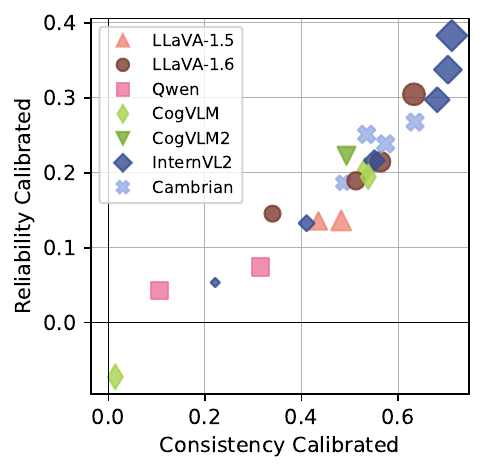}
         & \includegraphics[width=.24\linewidth]{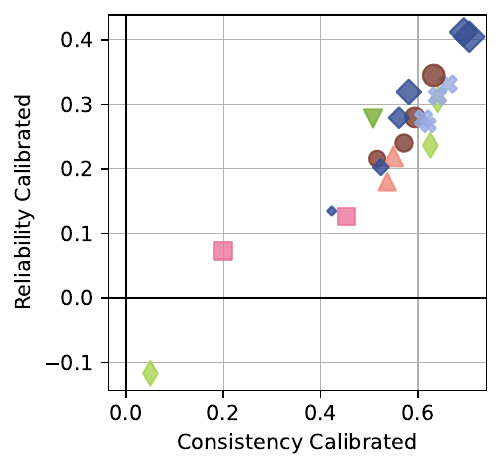}
         & \includegraphics[width=.24\linewidth]{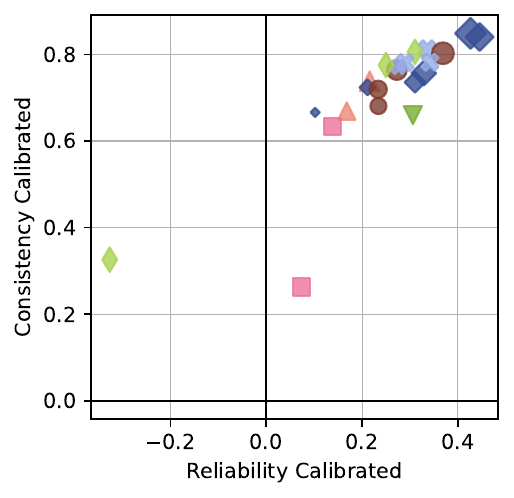}
         & \includegraphics[width=.24\linewidth]{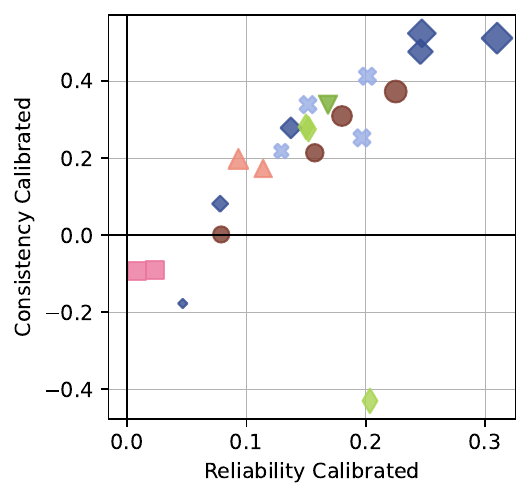}
   \end{tabular}
    \vspace{-0.3cm}
   \caption{Consistency vs. reliability over the four prompt variation types: Language, Vision, Reformulations and Semantic Changes. More detailed analysis to the overall averages in~\cref{tab:model-comp-reliab}. The top row shows the full ranges for calibrated reliability and consistency (from -1 to 1). The lower row shows zoom-ins. Marker size indicates VLM size. Model family rankings are stable across datasets. The most robust model family to prompt changes is InternVL2, with 40B as best.
   }
   \label{fig:supp_model-winner_fourvariations}
   \vspace{-0cm}
\end{figure*}

Below, we describe additional evaluation analysis for VLM prompt sensitivity. The first set of result is centered around expanding on~\cref{tab:model-comp-reliab}.
Then, we expand our arguments why data is most important for VLM prompt sensitivity.

\para{VLM sensitivity across datasets}
We also report detailed accuracy and certainty measures for our VLM sensitivity analysis in~\cref{tab:supp-model-comp-reliab}, expanding results from~\cref{tab:model-comp-reliab}. Again, accuracy and certainty trends follow our reliability score.

\begin{figure*}
   \centering
   \begin{tabular}{@{}c@{\ \ }c@{}c@{\quad}c@{}c@{}}
        & \multicolumn{2}{c}{\footnotesize\hspace{.5cm}\textbf{Reliability}} & \multicolumn{2}{c}{\footnotesize\hspace{.5cm}\textbf{Consistency}} \\
        \cmidrule(l{.8cm}r{.5cm}){2-3} \cmidrule(l{.8cm}r{.2cm}){4-5}
        & \footnotesize\hspace{.5cm}\textbf{Model} Size & \footnotesize\hspace{.5cm}\textbf{LLM} Size & \footnotesize\hspace{.5cm}\textbf{Model} Size & \footnotesize\hspace{.5cm}\textbf{LLM} Size \\
        \multirow{1}{*}[6.3em]{\rotatebox[origin=c]{90}{\footnotesize\textbf{All} Variations}}
        & \includegraphics[width=.24\linewidth]{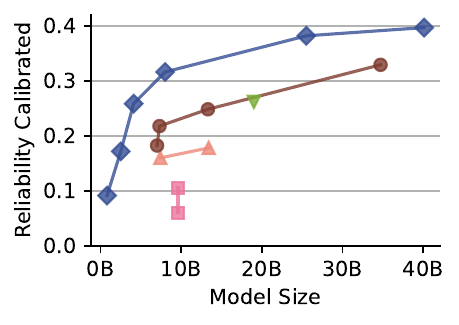}
         & \includegraphics[width=.24\linewidth]{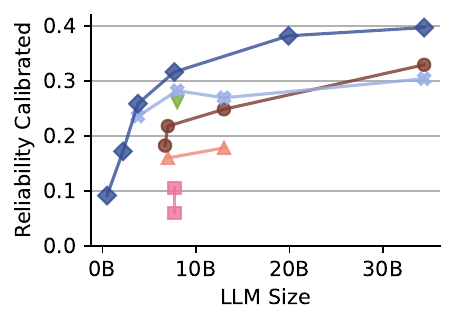}
         & \includegraphics[width=.24\linewidth]{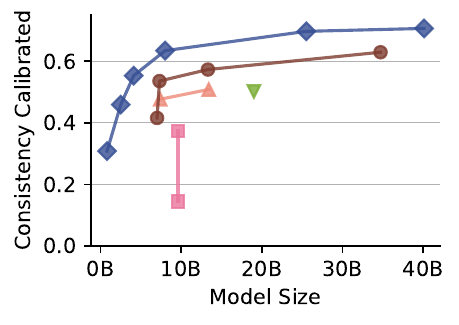}
         & \includegraphics[width=.24\linewidth]{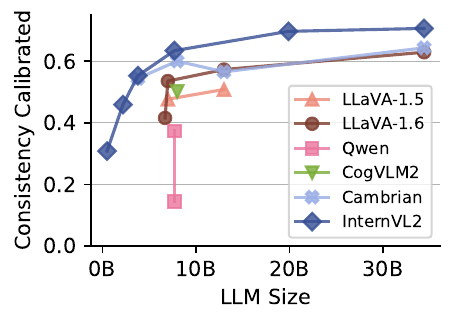}
         \\
        \multirow{1}{*}[7.5em]{\rotatebox[origin=c]{90}{\footnotesize\textbf{Language} Variations}}
        & \includegraphics[width=.24\linewidth]{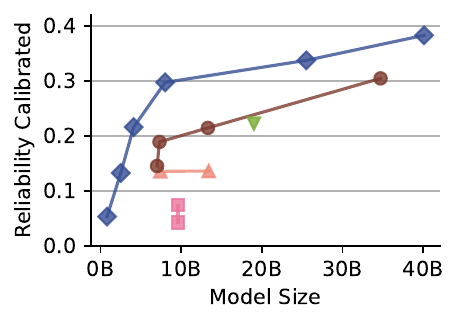}
         & \includegraphics[width=.24\linewidth]{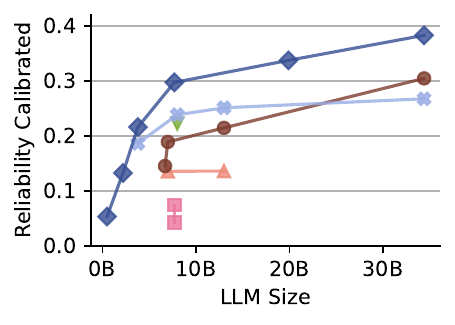}
         & \includegraphics[width=.24\linewidth]{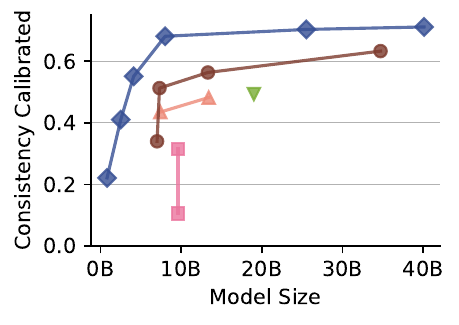}
         & \includegraphics[width=.24\linewidth]{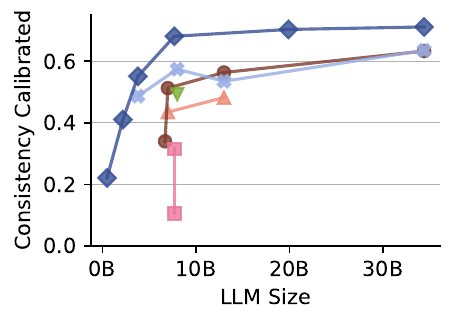}
         \\
        \multirow{1}{*}[6.8em]{\rotatebox[origin=c]{90}{\footnotesize\textbf{Vision} Variations}}
        & \includegraphics[width=.24\linewidth]{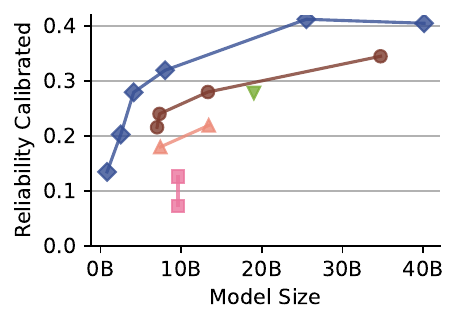}
         & \includegraphics[width=.24\linewidth]{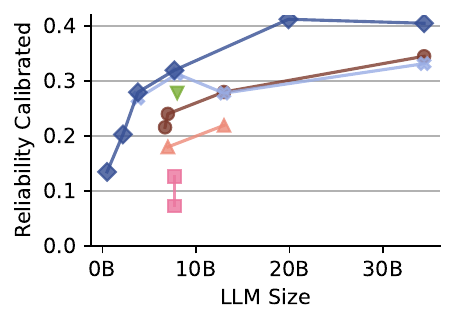}
         & \includegraphics[width=.24\linewidth]{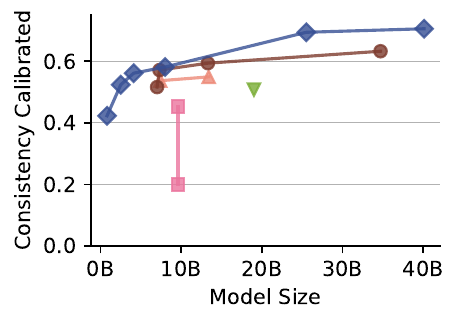}
         & \includegraphics[width=.24\linewidth]{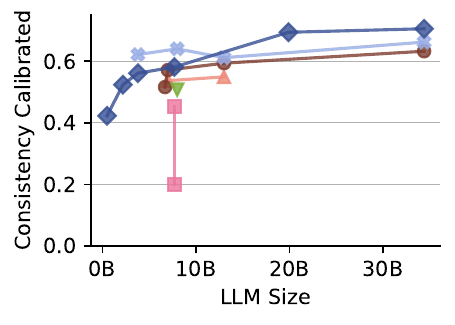}
   \end{tabular}
    \vspace{-0.3cm}
   \caption{Influence of Model and VLM size on reliability and consistency, extension of~\cref{fig:model-size}. The first row shows averages across all prompt variations, while the second and third row show averages over only language and only vision variations. The trends do not change when only one prompt modality is varied. Also, the trends across reliability and consistency remain comparable.
   }
   \label{fig:supp_model-size}
   \vspace{-.2cm}
\end{figure*}

\para{VLM sensitivity via logit evaluation}
We can also evaluate VLMs by extracting the logit values for all potential multiple choice answers, \ie the logits for \texttt{A}, \texttt{B}, \texttt{C} and \texttt{D}, and treat the highest as model answer.
This simplifies parsing the VLMs response and matching it to the possible answers, but may also generate false positive answers if the VLM generates a free form answer that does not match the answer choices at all.
We report the results in \cref{tab:supp-model-comp-reliab_logits}, which replicates~\cref{tab:model-comp-reliab} of the main paper with its extended results in~\cref{tab:supp-model-comp-reliab}.
Overall, the trends are well aligned.
The only major difference is CogVLM GG, which performs in line with the other CogVLM variants and does not underperform any longer if evaluated by logits.
This indicates that CogVLM GG generates text that does not resemble any answer choice but would find the correct answer if the response outputs were limited to the answer options.

\para{VLM sensitivity across variation classes}
To complement the right figure in~\cref{tab:model-comp-reliab} of the main paper, we show the consistency vs. reliability plots for our four main prompt variation types: Language variations, Vision variations, Reformulations and Semantic Changes in~\cref{fig:supp_model-winner_fourvariations}.
Across all variations, the ranking of methods remain very similar, and model families continue to be the main influence in VLM performance.
InternVL2 40B is still the most robust model, and within model families larger models also have an increased resilience against prompt variations.
The detailed analysis further reinforces our findings on prompt variation types: Reformulations are easier for VLMs to understand, and excluding semantic changes from sensitivity evaluations gives the wrong appearance of an overall improved prompt-agnosticism.

\para{VLM sensitivity across VLM and LLM sizes}
To complement the analyses in~\cref{fig:model-size}, we show reliability and consistency measurements over model- and LLM size in~\cref{fig:supp_model-size}.
In contrast to the main paper, we also include the reliability and consistency measurements over \textit{only} language and \textit{only} vision variations. Interestingly, the trends are very well aligned across language and vision variations, which further reinforces that architectural differences in the vision and language processing blocks do not sufficiently explain the differences in VLM prompt sensitivity.

{
    \small
    \bibliographystyle{ieeenat_fullname}
    \bibliography{main}
}

\end{document}